\documentclass[runningheads]{llncs}

\usepackage{ECCV-kit/eccv}

\usepackage{ECCV-kit/eccvabbrv}

\usepackage{graphicx}
\usepackage{booktabs}

\usepackage[accsupp]{axessibility}  

\usepackage{hyperref}

\usepackage{orcidlink}

\usepackage{amssymb}
\usepackage{subcaption}
\usepackage{pifont} 
\usepackage[ruled,linesnumbered]{algorithm2e}
\usepackage{bbm}
\usepackage{listings}

\newcommand{\OurMethod}{FreeAugment}

\newcommand{\AugmenterParams}{\phi}
\newcommand{\ClassifierParams}{\theta}

\newcommand{\SingleImage}{X}

\newcommand{\SingleOp}{\tau}

\newcommand{\MagUpperParam}{h}
\newcommand{\MagLowerParam}{l}

\newcommand{\temperature}{t}

\newcommand{\cmark}{\ding{51}}
\newcommand{\xmark}{\ding{55}}

\definecolor{darkgreen}{rgb}{0.1, 0.5, 0.1}
\definecolor{tiger}{rgb}{0.988,0.414,0.007}

\definecolor{codegreen}{rgb}{0,0.6,0}
\definecolor{codegray}{rgb}{0.5,0.5,0.5}
\definecolor{codepurple}{rgb}{0.58,0,0.82}
\definecolor{backcolour}{rgb}{0.95,0.95,0.92}

\lstdefinestyle{mystyle}{
    backgroundcolor=\color{backcolour},   
    commentstyle=\color{codegreen},
    keywordstyle=\color{magenta},
    numberstyle=\tiny\color{codegray},
    stringstyle=\color{codepurple},
    basicstyle=\ttfamily\footnotesize,
    breakatwhitespace=false,         
    breaklines=true,                 
    captionpos=b,                    
    keepspaces=true,                 
    numbers=left,                    
    numbersep=5pt,                  
    showspaces=false,                
    showstringspaces=false,
    showtabs=false,                  
    tabsize=2
}

\begin{document}

\title{FreeAugment: Data Augmentation Search \\ Across All Degrees of Freedom}

\titlerunning{\OurMethod}

\author{Tom Bekor$^*$\orcidlink{0009-0002-6660-7158}\and
Niv Nayman$^*$\orcidlink{0000-0001-6521-0090}\and
Lihi Zelnik-Manor\orcidlink{0000-0002-7930-8985}}
\authorrunning{T.~Bekor et al.}

\institute{Technion - Israel Institute of Technology, Haifa, Israel
\\ Corresponding author:
\email{tom.bekor@campus.technion.ac.il}}

\maketitle

\begin{abstract}
  Data augmentation has become an integral part of deep learning, as it is known to improve the generalization capabilities of neural networks. 
  Since the most effective set of image transformations differs between tasks and domains, automatic data augmentation search aims to alleviate the extreme burden of manually finding the optimal image transformations. However, current methods are not able to jointly optimize all degrees of freedom: (1) the number of transformations to be applied, their (2) types, (3) order, and (4) magnitudes.
  Many existing methods risk picking the same transformation more than once, limit the search to two transformations only, or search for the number of transformations exhaustively or iteratively in a myopic manner.
  Our approach, \OurMethod, is the first to achieve global optimization of all four degrees of freedom simultaneously, using a fully differentiable method.
  It efficiently learns the number of transformations and a probability distribution over their permutations, inherently refraining from redundant repetition while sampling.
  Our experiments demonstrate that this joint learning of all degrees of freedom significantly improves performance, achieving state-of-the-art results on various natural image benchmarks and beyond across other domains.\footnote{Project page:~\href{https://tombekor.github.io/FreeAugment-web}{https://tombekor.github.io/FreeAugment-web}}

\def\thefootnote{*}\footnotetext{Equal contribution.}\def\thefootnote{\arabic{footnote}}
  
  \keywords{Data 
  Augmentation \and AutoML \and Differentiable Optimization}
\end{abstract}

\section{Introduction}
\label{sec:intro}

\begin{figure}[htb]
	\centering
	\includegraphics[width=0.95\linewidth]{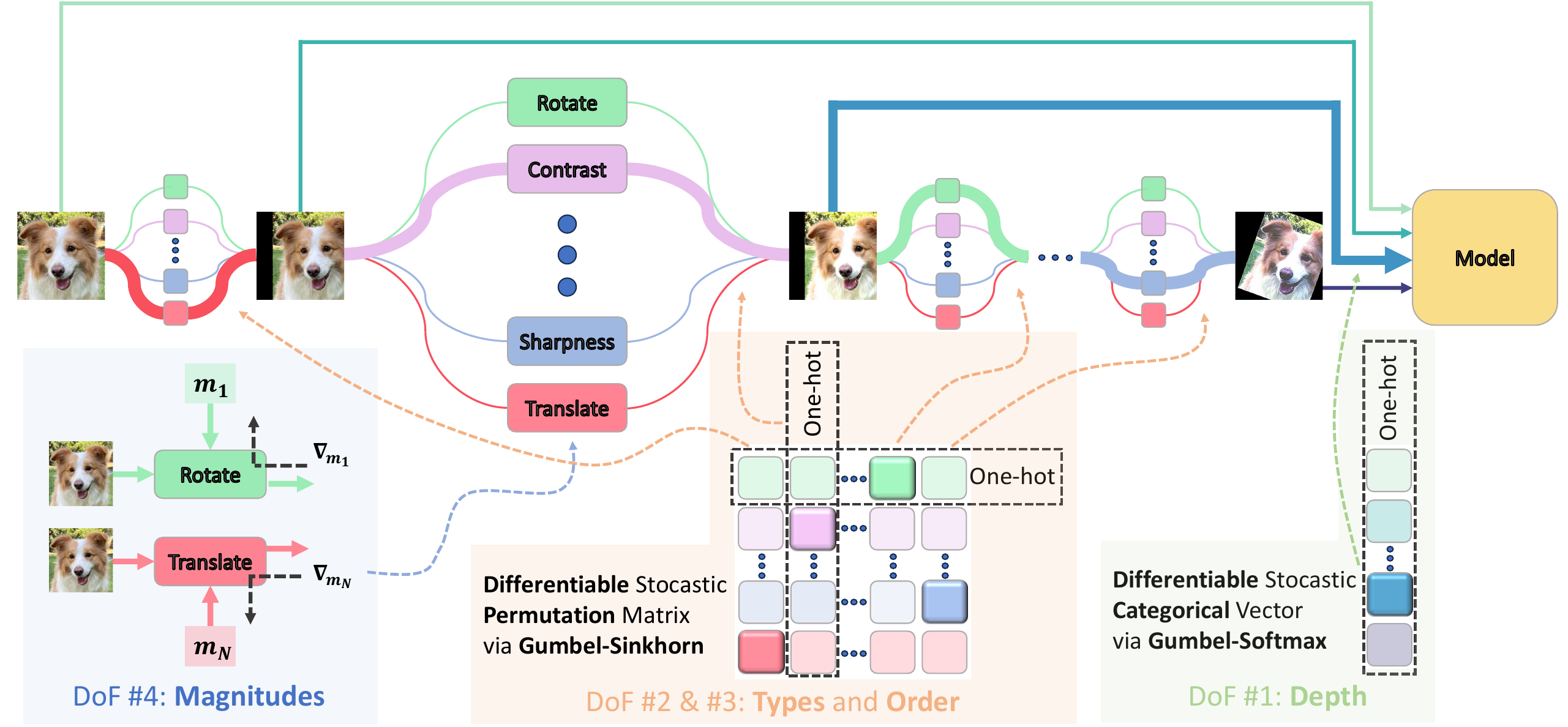}
	\caption{\OurMethod~differentiable search space over four degrees of freedom (DoF). The search space over policy depth, transformations type, order, and magnitudes is constructed in a fully differentiable manner and these are learned end-to-end.}
	\label{fig:policy_scheme}
 \vspace{-6mm}
\end{figure}

Data augmentations expand the training set by generating virtual samples through random transformations to the original ones.
This mitigates the overfitting \cite{shorten2019survey} problem when training large deep neural networks. However, designing a data augmentation pipeline for a given task requires domain knowledge, as different domains benefit from different types of transformations of different strengths.
This requirement has been alleviated by Data Augmentation Search (DAS), which enables the automatic customization of data augmentation for different tasks and domains. 
DAS methods sample from an optimized data augmentation policy, that consists of the probability distribution of choosing transformations from a predefined set together with their magnitudes (intensities). This combinatorial search space of policies makes the optimization problem challenging.

Differentiable DAS methods \cite{DADA, OnlineAugment, DABO, DifferentiableRandAugment, DDAS} have taken a gradient-based approach to solve a bilevel optimization problem that targets the best augmentation policy, such that a neural network trained with this policy on the training set (inner problem) generalizes well to a validation set (outer problem). 
This approach has shown good results; however, previous methods either limit the number of sequentially applied transformations (policy depth) to two~\cite{AutoAugment, PBA, FastAA, FasterAA, DADA, RandAugment, DDAS}, exhaustively enumerate through several policy depth values \cite{DifferentiableRandAugment, Slack}, or myopically increase the depth at every search iteration \cite{DeepAA}. In addition, seeking the best order of transformations, some methods sample predefined sub-policies, including all possible permutations of transformations \cite{AutoAugment, FastAA, FasterAA, DADA}. As the number of those grows exponentially with the policy depth, these methods limit the depth to at most two. Other methods sequentially sample transformations from independent categorical distributions and risk sampling the same transformation more than once~\cite{Slack, DifferentiableRandAugment, DeepAA, DDAS}.

In this paper, we aim to globally optimize the augmentation policy across all four degrees of freedom (DoF): (1) the number of sequentially applied transformations (policy depth), their (2) type, (3) order, and (4) magnitudes. 
To this end, we introduce \OurMethod, the first fully differentiable DAS method to jointly optimize all DoF in an end-to-end manner using gradient descent, without turning to complementary heuristics or greedy strategies. 
As illustrated in \cref{fig:policy_scheme}, \OurMethod~achieves all of that by (1) inducing a learnable probability distribution over the policy depth and thus can sample a varying depth. (2) Optimizing the distributions over types and order of transformations is unified as learning the probability distribution over permutations. This formalism inherently guarantees that images do not undergo any transformation more than once. In practice, this distribution is closely approximated by the Gumbel-Sinkhorn method~\cite{Gumbel-Sinkhorn}, which allows sampling from all possible permutations of a pre-defined set of transformations and learning the underlying probability distribution in a fully differentiable manner. 
Lastly, (3) stochastic magnitudes for each augmentation are sampled from a parameterized probability distribution, whose parameters are learned by backpropagation through the differentiable implementation of the candidate transformations during the search. This entire composition is optimized on a validation set in an end-to-end manner by bilevel optimization.

Our experiments show that the proposed framework is general and robust to a variety of domains.
It finds policies that lead to comparable or better results than other methods on the natural images benchmarks CIFAR10/100~\cite{CIFAR} and ImageNet-100~\cite{ImageNet-100}, as well as on DomainNet~\cite{DomainNet}, which includes sketches, paintings, drawings, cliparts and infographs.

To summarize, our key contributions are three-fold:
\begin{itemize}
    \item We propose a novel way to jointly learn the number of transformations in an end-to-end manner, without reverting to heuristics or multi-stage solutions.
    \item To avoid redundant repetition of transformations, we formalize the search over their types and order as learning the probability distribution over permutations of transformations, and utilize efficient tools for doing so in a fully differentiable manner.
    \item Through extensive experimentation, we showcase the synergistic impact of jointly learning all degrees of freedom of data augmentation policies. This leads to state-of-the-art performance on benchmarks of natural images and other domains compared to previous DAS methods.
\end{itemize}

\section{Related Work}
\label{sec:related_work}

To alleviate the domain-knowledge and manual labor required for choosing the right data augmentation policy for any given domain, AutoAugment (AA)~\cite{AutoAugment} was the pioneering work to automate the search of data augmentations. This was done by modeling the policy search problem as a sequence prediction problem, and using a recurrent neural network controller in a Reinforcement Learning (RL) framework to predict the type and magnitude of at most two sequential transformations. 
Since RL tends to suffer from high variance, this method is sample inefficient and hence requires a large amount of data and compute. 
A following evolutionary approach, PBA~\cite{PBA}, is based on sampling population of policies and hence still expensive. A Bayesian optimization~\cite{pmlr-v216-tan23a} based method, FastAA\cite{FastAA} does not leverage gradient information, and thus in practice those all are not learned in an end to end manner and do not perform better than random search, e.g., RandAugment~ \cite{RandAugment} and TrivialAugment~\cite{TrivialAugment}.
Inspired by differentiable neural architecture search \cite{liu2018darts, nayman2019xnas, pmlr-v108-noy20a, pmlr-v189-nayman23a}, differentiable DAS formulate the search space of augmentations so that it can be optimized using gradient decent. FasterAA~\cite{FasterAA} minimizes the distance between the original and augmented image distributions, AdversarialAA~\cite{AdversarialAA} and TeachAugment~\cite{TeachAugment} maximize the training loss with respect to the augmentation policy under some regularizations, and MADAO~\cite{hataya2022meta} minimizes the validation loss directly with Neumann series approximation of the gradients.
MedPipe~\cite{MedPipe}, DAAS~\cite{DAAS} and DHA~\cite{DHA} jointly search for augmentation policies and network architectures.
OHLA~\cite{OHLA}, DADA~\cite{DADA}, DDAS~\cite{DDAS}, DABO~\cite{DABO},  DRA~\cite{DifferentiableRandAugment} and SLACK~\cite{Slack} use meta-learning with one-step gradient update and differentiable optimization. 
Most of the aforementioned \cite{AutoAugment, PBA, RandAugment, TrivialAugment, FastAA, AdversarialAA, hataya2022meta, OHLA, DADA, DDAS} limit the number of sequential transformations to two, and some of those \cite{FasterAA, DifferentiableRandAugment} further enumerate several possible values for this number (up to four transformations) but eventually fixate on a couple of transformations as well. DeepAA~\cite{DeepAA} was the first to search beyond two transformations, but it does so by myopically stacking augmentation layers one at a time until convergence to an identity layer. \OurMethod~is the first to perform global end-to-end optimization over the distribution of the number of transformations in a fully differentiable manner. Additionally, all of the listed methods, but DeepAA and SLACK, determine the order of applied transformations by sampling from predefined sub-policies of all possible permutations. The number of those grows exponentially with the policy depth, which is hence limited to be at most two. \OurMethod~is the first to learn the distribution over all permutations of different lengths.

\section{Method}

\label{sec:SearchSpace}

Data augmentation can be viewed as a sequential application of image transformations over an input image $\SingleImage$, where an image transformation $\SingleOp$ with an application \emph{magnitude} $m$ transforms the input image into another \emph{augmented} image $\SingleOp(\SingleImage; m)$ of the same size. This magnitude defines how strongly to transform the input image, e.g., the rotation angle for a rotation transformation.
Image transformations are selected from a predefined set $\{\SingleOp_1,\dots,\SingleOp_N\}$ of $N$ candidate \emph{elementary transformations}.

Following previous work~\cite{Slack, DADA, FastAA, FasterAA, AutoAugment}, we define the augmentation policy as a probabilistic model 
$\mathcal{P}_{\AugmenterParams}$
that generates a sequence of elementary transformations with corresponding magnitudes, governed by parameters $\AugmenterParams$ as outlined in \cref{alg:policy-sampling}. Those parameters are divided into three groups, $\delta$, $\Pi$, and $\mu$, associated with the different degrees of freedom. 

\RestyleAlgo{ruled}
\SetKwComment{Comment}{// }{}
\vspace{-2mm}
\begin{algorithm}[hbt]
\caption{\OurMethod~Policy Sampling at Search}\label{alg:policy-sampling}
\KwData{Input image $\SingleImage_0$, policy parameters $\AugmenterParams = (\delta, \Pi, \mu )$}
\KwResult{Augmented image $\SingleImage'$}
$ \mathbf{d} \sim \text{Gumbel-Softmax}(\delta)$ \Comment*[r]{Sample depth one-hot vector (\cref{subsec: Augmentations Depth})}
$ \mathbf{P} \sim \text{Gumbel-Sinkhorn}(\Pi)$ \Comment*[r]{Sample a permutation matrix (\cref{subsec: Augmentations Order})}
$ \mathbf{M} \sim \text{Uniform}(\mu)$ \Comment*[r]{Sample a magnitude matrix (\cref{subsec: Augmentation Magnitude})}
\For (\Comment*[f]{Iterate over augmentation layers}) {$k \in \{1,\dots,K\}$}
{   
        $ \SingleImage_k \gets \sum_{i=1}^N \mathbf{P}_{ik} \cdot \SingleOp_i (\SingleImage_{k-1},\mathbf{M}_{ik})$
}
$ \SingleImage' \gets \sum_{k=0}^K \mathbf{d}_k \cdot \SingleImage_k$ 
\end{algorithm}
\begin{figure}[htb]
    \centering
    \begin{subfigure}{0.65\textwidth}
        \includegraphics[width=\textwidth]{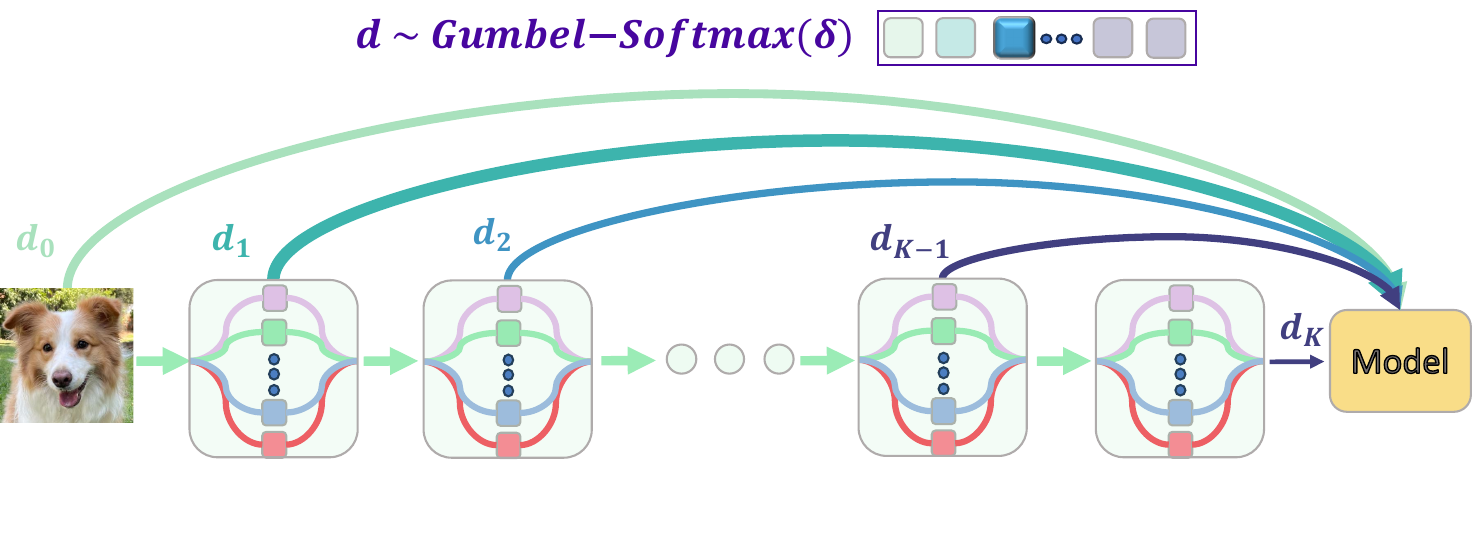}
    \end{subfigure}
    \begin{subfigure}{0.33\textwidth}
        \includegraphics[width=\textwidth]{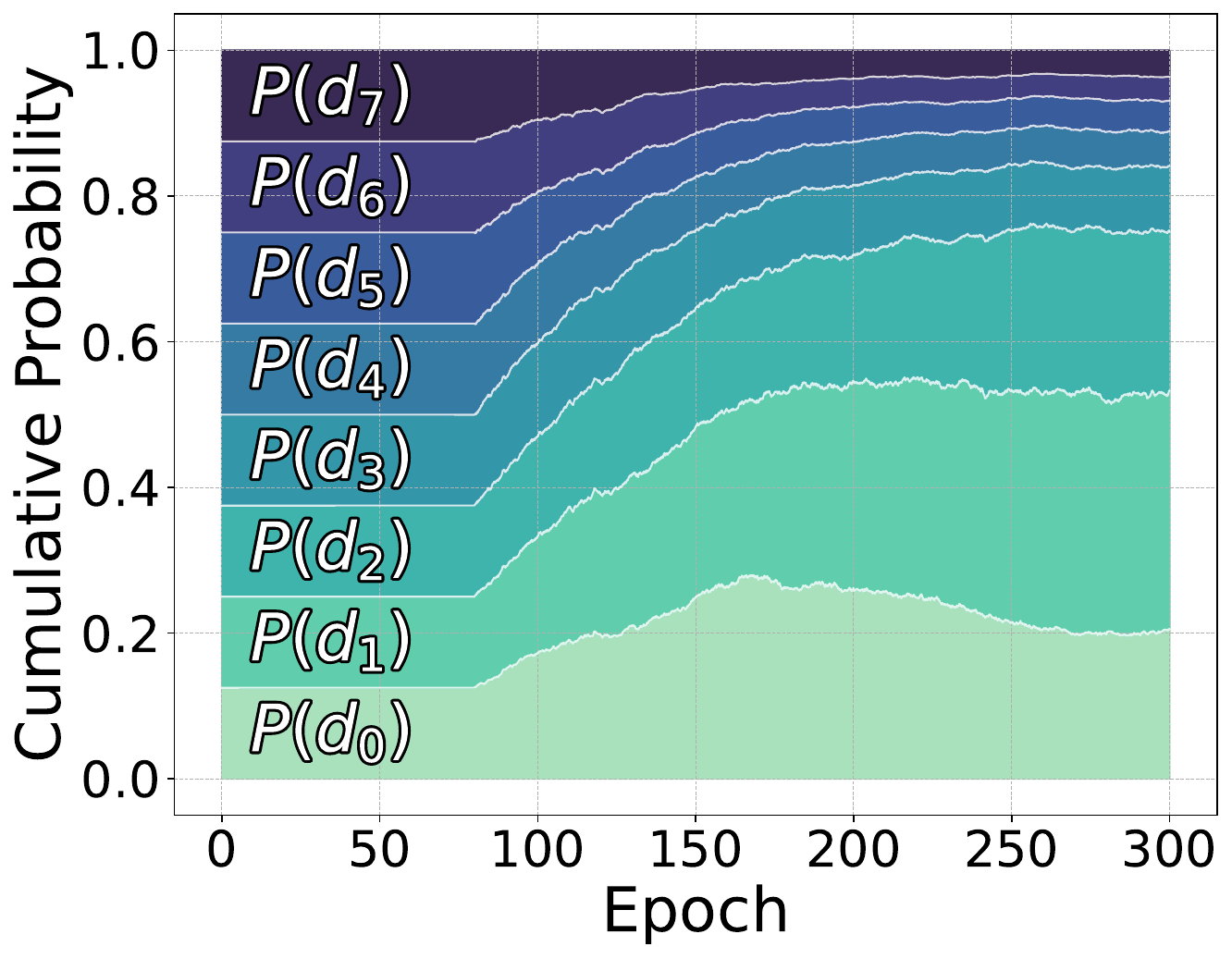}
    \end{subfigure}
    \caption{Illustration of \OurMethod's policy depth search mechanism (Left) and learned depth distribution along the search (Right), where the last epoch represents the resulting probability distribution over policy depth.}
    \label{fig:depth}
 \vspace{-1mm}
\end{figure}

\subsection{$1^{\text{st}}$ Degree of Freedom: Depth}
\label{subsec: Augmentations Depth}

The policy depth refers to the number of sequential transformations applied to a single image. 
To dynamically determine this depth, inspired by \cite{hu2020tf, pmlr-v139-nayman21a}, we induce a learnable Gumbel-Softmax distribution \cite{Gumbel-Softmax} over categorical depths. It is parameterized by a vector of logits $\delta$, where each element $\delta_k$ for $k\in\{0,\dots,K\}$ represents the unnormalized log-probability of selecting an augmentation policy with depth $k$, as illustrated in \cref{fig:depth} (Left). 
This results in a probability distribution over the depths that is smooth and differentiable:
\begin{equation}
\mathbf{d} \sim \text{Gumbel-Softmax}(\delta; \temperature),\quad\text{such that } \mathbb{P}(d_k=1 \mid \temperature) = \frac{e^{(\delta_k + g_k) / \temperature}}{\sum_{i=1}^{K} e^{(\delta_i + g_i) / \temperature}}
\end{equation}

where $g_i$ are i.i.d samples from a standard Gumbel distribution, and $\temperature$ is the temperature parameter.
Effectively, the straight-through (ST) gradient estimator \cite{straight-through-estimator} is utilized to to efficiently learn the logits $\delta$ by backpropagating through the discrete choice of augmentation depth of maximal value in $\mathbf{d}$, as demonstrated in \cref{fig:depth} (Right).

\subsection{$2^{\text{nd}}$ and $3^{\text{rd}}$ Degrees of Freedom: Types \& Order}
\label{subsec: Augmentations Order}

\begin{figure}[htb]
	\centering
	\includegraphics[width=1.0\linewidth]{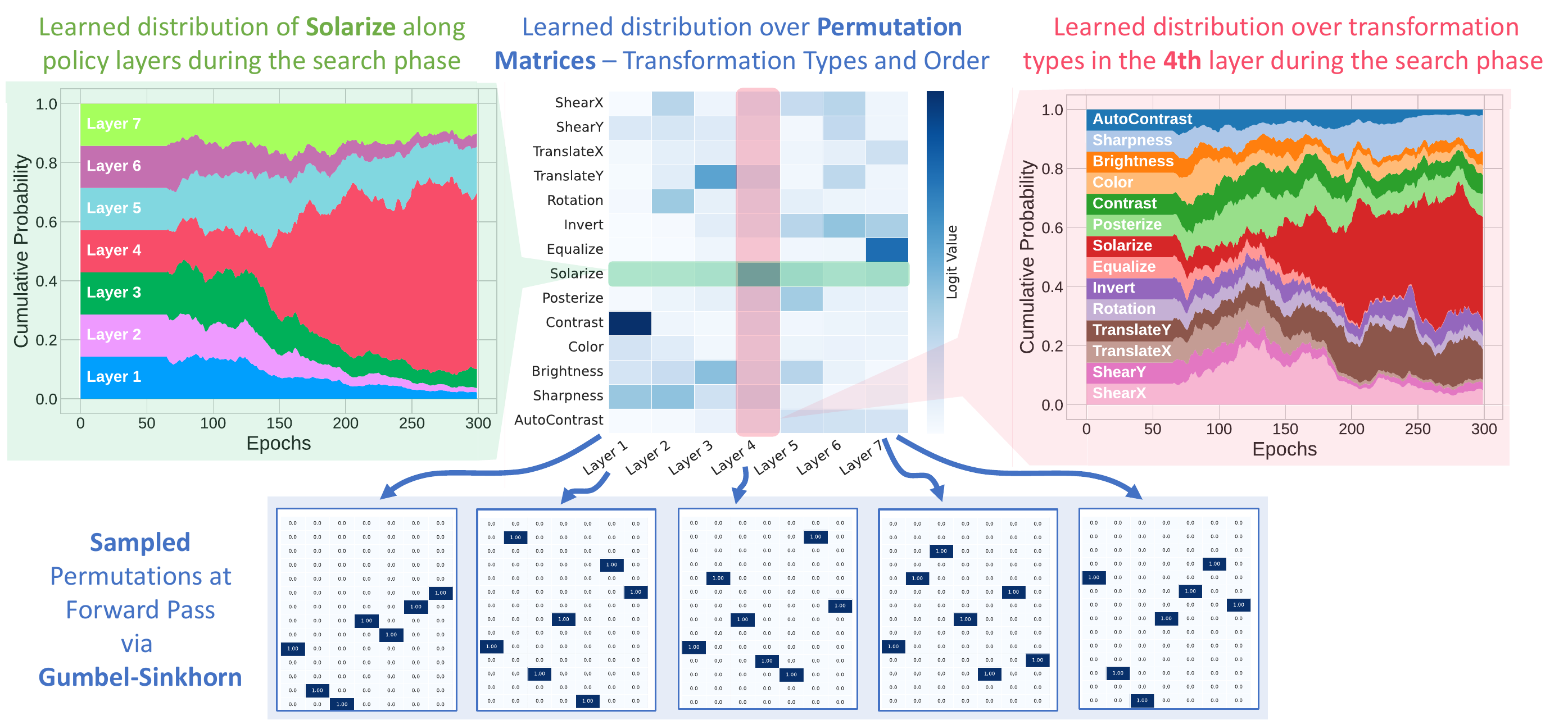}
	\caption{Learning the transformation types and order as the distribution over permutations of transformations (Top Middle) and sampling from it (Bottom). The learning dynamics of a column (Top Right) and a row (Top Left) of the matrix along the search are shown, with the final values embedded in the matrix.}
    \label{subfig:permutation_dynamics}
    \vspace{-5mm}
\end{figure}

Learning the distribution over types and order of transformations, requires sampling a transformation for every augmentation layer. Doing so for each layer independently using the Gumbel-Softmax distribution, as in \cite{DifferentiableRandAugment}, permits the repetition of the same transformations in several layers.

To mitigate that, we view the search for the types and order of transformations as a matching problem, where we seek to match between $N$ elementary transformations to $K$ augmentation layers. Thus, we couple the different layers and learn a probability distribution for sampling an entire permutation matrix $\mathbf{P}$ of size $N\times K$, where not only its columns are normalized but also its rows. Its $k^{th}$ column $\mathbf{P}_{1:N,k}$ represents the distribution over transformations at augmentation layer $k$, as illustrated the middle of \cref{fig:policy_scheme} and the top middle of \cref{subfig:permutation_dynamics}.

To this end, we utilize the Gumbel-Sinkhorn operator~\cite{Gumbel-Sinkhorn} for differentiating through permutation sampling as an analogy of the Gumbel-Softmax for differentiating through categorical sampling.

We first state an analog of the normalization implemented by the softmax; a double stochastic matrix (DSM) is a square matrix of non-negative real numbers with the property that every row and every column sums to one. 
Sinkhorn \cite{sinkhorn1964relationship, sinkhorn1967concerning} showed that any non-negative square matrix $\exp(\bar{\Pi})\in\mathbb{R}^{N\times N}$ (with full support \cite{knight2008sinkhorn}) can be converted to a DSM by alternating between rescaling its rows and columns to one using the Sinkhorn operator~\cite{adams2011ranking, pmlr-v139-indelman21a}: 
\begin{equation}
    \label{eq:sinkop} 
     S(\bar{\Pi}) = \lim_{l\rightarrow \infty} S^l(\bar{\Pi}) ;
    \quad
     S^l(\bar{\Pi}) = \mathcal{N}_c\left(\mathcal{N}_r(S^{l-1}(\bar{\Pi}))\right); 
     \quad 
    S^0(\bar{\Pi}) = \exp(\bar{\Pi})
\end{equation}
where $\mathcal{N}_r(\bar{\Pi})= \bar{\Pi} \oslash (\bar{\Pi} \mathbf{1}_N\mathbf{1}_N^\top),$ and $\mathcal{N}_c(\bar{\Pi})= \bar{\Pi}   \oslash (\mathbf{1}_N\mathbf{1}_N^\top \bar{\Pi})$ as the row and column-wise normalization operators of a matrix, with $\oslash$ denoting the element-wise division and $\mathbf{1}_N$ a column vector of ones. 

Next we move from the generation of DSMs to sampling from permutation matrices and learn the types and order of transformations. According to the Birkhoff-von Neumann theorem \cite{birkhoff1946three, von1953certain}, the Birkhoff polytope $\mathcal{B}_N$ (which is the set of $N\times N$ DSMs), forms a convex hull for the set of $N\times N$ permutation matrices. Consequently, it is natural to think of DSMs as relaxations of permutation matrices. 
However, we aim at learning a $N\times K$ matrix of real logits $\Pi\in\mathbb{R}^{N\times K}$ to induce a distribution over permutation matrices of the same size, representing the assignment of N types of transformations to K layers. This matrix is not square and neither non-negative. Inspired by \cite{wang2020combinatorial, wang2020graduated, cho2010reweighted, yew2020rpm, wang2021neural}, we bypass that by padding $\Pi$ with additional $N - K$ columns occupied by extremely negative constant values, turning it into a square matrix $\Bar{\Pi}$ of size $N\times N$, whose exponent is taken in \cref{eq:sinkop} to achieve non-negativity.

In practice, sampling from the Gumbel-Sinkhorn distribution of DSMs is done by applying the Sinkhorn operation for $L$ iterations over a perturbed padded matrix and truncating the padded part, 
\begin{equation}
    \mathbf{P} = \mathbf{\bar{P}}_{1:N,1:K}\quad ; 
    \quad
    \mathbf{\bar{P}} = S^L((\bar{\Pi}+G) / \temperature)
\end{equation}
where $G$ is a matrix of standard i.i.d. Gumbel noise of size $N\times N$.

Note that those samples are merely close approximations of permutation matrices. 
As proven by \cite{Gumbel-Sinkhorn}, as the temperature $\temperature$ decreases and the number of Sinkhorn iterations $L$ increases, the resulting matrix $\mathbf{P}$ gets closer to a binary permutation matrix (\cref{subfig:permutation_dynamics} top middle). However, this comes at the cost of vanishing gradients due to low temperature and more compute due to the many iterations \cite{Gumbel-Sinkhorn}. To refrain from that during the search phase, the fine corrections required for conversion to binary permutation matrices are done by selecting the index of maximal value at every column of $\mathbf{P}$ (\cref{subfig:permutation_dynamics} bottom). Again, the ST estimator \cite{straight-through-estimator} is utilized for learning the logits $\Pi$ by backpropagating through this discrete choice of transformation types and order, as demonstrated on the top right and top left of \cref{subfig:permutation_dynamics} respectively.

This effect is also present during the evaluation phase, as a finite choice of $L$ and non-zero temperature might result in deviations from a valid permutation matrix with all rows and columns being one-hot vectors. However, in practice this rarely happens, and the repetitions of the same transformations over different augmentation layers drops significantly, as shown in \cref{sssec:ablation-gumbel-sinkhorn-vs-softmax} and \cref{subfig:ablation-repetitive-sampling}.

\subsection{$4^{\text{th}}$ Degree of Freedom: Magnitudes}

\label{subsec: Augmentation Magnitude}

\begin{figure}[htb]
\vspace{-5mm}
	\centering
	\includegraphics[width=\linewidth]{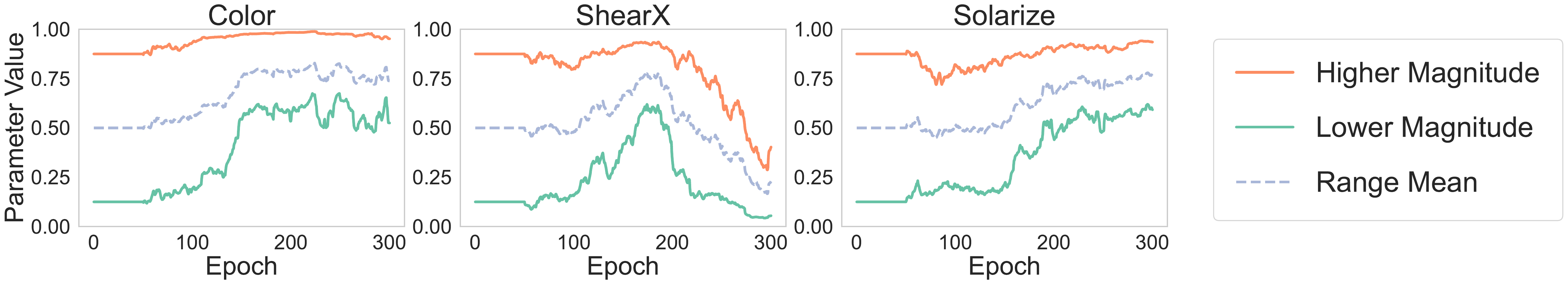}
	\caption{The dynamics of the learnable magnitude ranges of transformations at the first augmentation layer during the search over the CIFAR10 datasets with WRN-40-2. The lower and upper bounds of the uniform distribution are presented with solid lines and with their mean value with a dashed line.}
	\label{subfig:magnitudes}
\end{figure}

We follow previous studies \cite{Slack}, which found uniform sampling of transformation magnitudes to work on par with more elaborate sampling strategies, while their range has more impact on the results. 
Thus, for the elementary transformation $\tau_i$ of the $k^{th}$ augmentation layer we set $M_{ik}\sim \text{Uniform}(\mu_{ik})$ with the ranges $\mu_{ik}=(l_{ik}, u_{ik})$ consisting of the learnable lower and upper bounds $l_{ik}, u_{ik}\in\mathbb{R}$ respectively for $i\in\{1,\dots,N\}$ and $k\in\{1,\dots,K\}$.

In order to learn those magnitude ranges via backpropagation, as demonstrated in \cref{subfig:magnitudes}, we utilize a differentiable implementation of the elementary transformations \cite{Kornia} and the  reparameterization trick \cite{ReparameterizationTrick} to backpropagate through the samples of magnitudes, 
\begin{equation}
    \label{eq:magnitude}
    M_{ik} =  [\sigma(\MagUpperParam_{ik}) - \sigma(\MagLowerParam_{ik})] \cdot \epsilon + \sigma(\MagLowerParam_{ik}) 
\end{equation}
with $\epsilon\sim \text{Uniform}(0,1)$ and the sigmoid function $\sigma(\cdot)$ scaling the real learnable variables into the range $(0,1)$. We then scale $M_{ik}$ to the accepted input range by the corresponding transformation. 
Some transformations are not differentiable (e.g., posterize and solarize); thus, the ST estimator is used to evaluate the gradients of their magnitude; see the supplementary materials for more details.

\subsection{Policy Search as a Bilevel Optimization}
\label{subsec:bilevel pipeline}
\label{subsec: Bilevel Optimization}

Once we outlined a fully differentiable way to sample from the introduced search space of data augmentation policies via \cref{alg:policy-sampling}, we follow a large portion of differentiable DAS methods \cite{OHLA, DADA, DDAS, DABO, DifferentiableRandAugment, Slack} and learn the underlying parameters $\phi$ by solving a bilevel optimization problem,

\begin{align}
\label{eq:upper_problem}
&\min_\phi \mathcal{L}_{val}(\theta^*(\phi)) \\
\label{eq:lower_problem}
&\text{s.t.} \quad \theta^*(\phi) = \text{arg}\underset{\theta}min\mathcal{L}_{train}(\theta, \phi)
\end{align}
where,
\begin{align}
\mathcal{L}_{val}(\theta)&= \mathbb{E}_{(x,y)\sim\mathcal{D}_{val}}\mathcal{L}_{CE}\left(\mathcal{M}_\theta\left(x\right), y\right)  \\ 
\mathcal{L}_{train}(\theta, \phi)&= \mathbb{E}_{(x,y)\sim\mathcal{D}_{train}}\mathbb{E}_{\mathcal{T}\sim\mathcal{P}_\phi}\mathcal{L}_{CE}\left(\mathcal{M}_\theta(\mathcal{T}(x)), y\right)
\end{align}

and $D_{train}, D_{val}$ are the train and validation splits respectively, $\mathcal{L}_{CE}$ is the cross entropy loss, $\mathcal{M}_\theta$ is the model with parameters $\theta$, and $\mathcal{T}$ is an augmentation pipeline sampled from the augmentation policy $\mathcal{P}_\phi$ parameterized by $\phi$ according to \cref{alg:policy-sampling}.

Directly solving the upper problem in \cref{eq:upper_problem} would require solving the lower problem in \cref{eq:lower_problem} for every optimization step over the upper problem. To avoid that, we revert to the single step approximation and optimize $\ClassifierParams$ and $\AugmenterParams$ alternately through stochastic gradient descent, with the update terms,

\begin{align}
\Delta_{\theta_i}(\phi)&=\nabla_\theta\mathcal{L}_{train}(\theta_i, \phi) 
\label{eq:virtual step} \\
\Delta_{\phi_i} &= \nabla_{\phi}\mathcal{L}_{val}(\theta_i-\eta\Delta_{\theta_i}(\phi_i))
=
-\eta\nabla^2_{\theta, \phi}\mathcal{L}_{train}(\theta_i, \phi_i)  \nabla_\theta\mathcal{L}_{val}(\theta_i)
\label{eq: validation gradient} 
\end{align}

Thus $\theta_{i+1}$ and $\phi_{i+1}$ are updated by an optimizer of choice with $\Delta_{\theta_i}(\phi_i)$ and $\Delta_{\phi_i}$ respectively. 
In practice, we replace all expectations by estimates on a batch of data samples and sampled augmentations. 
The variance of those estimates with respect to the sampled augmentations is significantly reduced by sampling a distinct augmentation policy for each image in the batch rather than sampling once for the entire batch. This is enabled by an efficient implementation of \cref{alg:policy-sampling}, inspired by \cite{pmlr-v139-nayman21a} as detailed in the supplementary material. Furthermore, the computation of second order derivatives in \cref{eq: validation gradient} is reasonable since the size of $\phi$ is much smaller than the size of $\theta$, and thus it is performed efficiently using automatic differentiation software \cite{pytorch, TorchOpt}.

\section{Experiments}
Next we describe our experimental setup and evaluation process in \cref{subsec:experimental-setup}. In \cref{subsec:comparison-with-sota} our method is compared to previous work over a variety of common natural images benchmarks, as well as a variety of several other domains in \cref{subsec:domainnet}. Additionally, in \cref{subsec:ablations} we delve into an empirical analysis of insightful aspects of our approach including ablation studies of the main components of our method. This comprehensive evaluation showcases the effectiveness of \OurMethod~and provides insights into its properties.

\subsection{Experimental Setup}
\label{subsec:experimental-setup}
\subsubsection{Benchmarks}
We evaluate the performance of \OurMethod~on three standard benchmarks; CIFAR10, CIFAR100\cite{CIFAR}, and ImageNet-100\cite{ImageNet-100}, which are all natural images. We also evaluate on the DomainNet\cite{DomainNet} dataset, which consists of six distinct domains; real, quickdraw, infograph, sketch, painting, and clipart images, with a total of approximately 0.6 million images of 345 categories. Following \cite{Slack}, we use a reduced train set of 50k images for the two largest domains, real and quickdraw, and leave the rest of the samples for testing, according to the published file names of train and test images for each domain by \cite{Slack}. 
In alignment with previous work~\cite{AutoAugment, TrivialAugment, DeepAA, FastAA, DDAS, Slack}, Wide-ResNet40-2 and Wide-ResNet28-10 \cite{WideResNet} architectures are used for CIFAR10/100 datasets and ResNet18 for ImageNet-100 and DomainNet\cite{ResNet}.

\subsubsection{Baselines}
Following previous works \cite{AutoAugment, PBA, DeepAA}, we compare \OurMethod~to a baseline based on standard augmentations: random horizontal flipping and random pad-and-crop for CIFAR-10/100, an Inception-style preprocessing \cite{TIMM} for ImageNet-100, and a set of augmentations used by DomainBed \cite{DomainBed} for domain generalization on DomainNet datasets, including random horizontal flipping, color jitter (which is a mix of brightness, contrast, and saturation) and random gray-scale. Additionally, we compare \OurMethod~to many other existing methods, including:
 AA \cite{AutoAugment},
 PBA \cite{PBA},
 FastAA \cite{FastAA},
 FasterAA \cite{FasterAA},
 DADA \cite{DADA},
 DABO \cite{OHLA},
 RandAugment (RA) \cite{RandAugment},
 UniformAugment (UA) \cite{UniformAugmnet},
 TrivialAugment (TA) \cite{TrivialAugment},
 TeachAugment \cite{TeachAugment}
 MADAO \cite{hataya2022meta}
 DeepAA \cite{DeepAA},
 SLACK \cite{Slack},
 DRA \cite{DifferentiableRandAugment},
 and 
 DDAS \cite{DDAS}.

\subsubsection{Search Space}
\OurMethod's search space is designed to enable the joint optimization of augmentation policies across all 4 DoF depicted in \cref{fig:policy_scheme}.
In all of our experiments we search across augmentation policies with the maximal depth of $K=7$. 
We adopt AutoAugment's space of possible transformation types, except for SamplePairing\cite{SamplePairing} and Identity.
These include: ShearX, ShearY, TranslateX, TranslateY, Rotate, Solarize, Posterize, Contrast, Color, Brightness, Sharpness, AutoContrast, Invert and Equalize. 
All of which but the last three are associated with learnable magnitudes, whose ranges are detailed in the supplementary materials. 
Following conventional training practices and previous work~\cite{AutoAugment, PBA, TrivialAugment}, cropping, horizontal flipping, and cutout were applied by default.

\subsubsection{Policy Search} All search phases are conducted on reduced subsets of the data, each containing approximately 10\% random samples: 5,000 images for CIFAR10 and CIFAR100, 10,000 images for ImageNet-100, and 6,000 images for each of the six domains in DomainNet. 
The search phase is conducted on the reduced datasets, with a balanced 50\%-50\% train-validation split.
We search for 300 epochs, with a batch size of 128 and 64 for CIFAR10/100 and ImageNet-100/DomainNet, respectively. 

The policy distribution is initialized to be uniform over the search space. The magnitude parameters are initialized to $(0.125,0.875)$ in the normalized range of $(0,1)$. We use the Adam optimizer \cite{Adam} for all the policy's learnable parameters, and set different learning rates of 0.02, 0.01, 1.0 for the policy's magnitude, types, and depth learnable parameters.
To allow the model to learn before participating in the search, similarly to \cite{DeepAA,Slack}, we use warm-up to start its training before learning the policy. Since we jointly learn 4 DoF of different impact levels, gradual warm-up periods are applied for 50, 65, and 80 epochs, respectively, for magnitude, types, and depth. 
Throughout the search phase, Gumbel-based distributions are scaled by a temperature parameter that is exponentially annealed from 1.0 to 0.5, and the number of Sinkhorn iterations is set to $L=20$.
These search configurations are kept the same throughout all experiments and ablations, highlighting \OurMethod's robustness to hyperparameters without necessitating additional adjustments for different datasets and architectures.

\subsubsection{Policy Evaluation} We evaluated the performance of the generated data augmentation policies across the complete  CIFAR10, CIFAR100, ImageNet-100, and DomainNet datasets, replicating the exact training configurations in previous work for fair comparison. All hyperparameters are detailed in the supplementary materials to allow full reproducibility. During evaluation, the number of Sinkhorn iterations is set to $L=20$ and the temperature for all Gumbel-based distributions was consistently set to 0.1, as there is no longer a necessity for differentiation through these. The reported top-1 test accuracy (\%) for each experiment is averaged over 5 runs with different random seeds, with the 95\% confidence interval denoted by $\pm$.

\subsection{Comparison with Previous Approaches}
\label{subsec:comparison-with-sota}

\subsubsection{CIFAR10 and CIFAR100}
 Table \ref{tab:cifar_results} provides performance evaluation of augmentation policies generated by \OurMethod~on CIFAR10 and CIFAR100 for Wide-ResNet40-2 and Wide-ResNet28-10 compared to the results reported by other state-of-the-art methods.
 \OurMethod~outperforms the sota on CIFAR10 for both architectures and on CIFAR100 for Wide-ResNet40-2, with comparable performance to many other DAS methods. Specifically, \OurMethod~improves over the baseline of basic augmentations by $\textbf{+1.05\%}$, 
 $\textbf{+1.18\%}$, $\textbf{+3.3\%}$, $\textbf{+2.39\%}$ for Wide-ResNet40-2 and Wide-ResNet28-10 on CIFAR10 and CIFAR 100 respectively.

\begin{table}[h]
\centering
\caption{Top-1 test accuracy (\%) on CIFAR10/100 for WRN-40-2 and WRN-28-10 as reported by the original papers. Our results are averaged over 5 random seeds, with the 95\% confidence interval denoted by $\pm$.}
\label{tab:cifar_results}
\begin{tabular}{@{}l@{\hspace{4mm}}l@{\hspace{4mm}}l@{\hspace{4mm}}l@{\hspace{4mm}}l@{}}
\toprule
& \multicolumn{2}{c}{CIFAR10} & \multicolumn{2}{c}{CIFAR100} \\
\cmidrule(r){2-3} \cmidrule(){4-5}
& WRN-40-2 & WRN-28-10 & WRN-40-2 & WRN-28-10 \\
\midrule
Baseline & 95.49 $\pm$ .14 & 96.48 $\pm$ .15 & 76.74 $\pm$ .29 & 81.75 $\pm$ .12 \\
\midrule
AA\cite{AutoAugment} & 96.3 & 97.4 & 79.3 & 82.9 \\
PBA\cite{PBA} & - & 97.42 $\pm$ .06 & - & 83.27 $\pm$ .15 \\ 
FastAA\cite{FastAA} & 96.4 & 97.3 & 79.4 & 82.7 \\
DADA\cite{DADA} & 96.4 & 97.3 & 79.1 & 82.5 \\
RA\cite{RandAugment} & - & 97.3 & - & 83.3 \\
TeachA\cite{TeachAugment} & - & 97.5 & - & 83.2 \\
UA\cite{UniformAugmnet} & 96.25 & 97.33 & 79.01 & 82.82 \\
TA(Wide)\cite{TrivialAugment} & 96.32 $\pm$ .05 & 97.46 $\pm$ .06 & 79.86 $\pm$ .19 & \textbf{84.33 $\pm$ .17} \\
FasterAA\cite{FasterAA} & 96.30 & 97.40 & 78.60 & 82.70 \\ 
DABO\cite{DABO} & - & 96.44 & - & 81.90 \\
DeepAA\cite{DeepAA} & - & 97.56 $\pm$ .14 & - & 84.02 $\pm$ .18 \\
DRA\cite{DifferentiableRandAugment} & - & 97.44 $\pm$ .10 & - & 83.85 $\pm$ .16 \\ 
DDAS\cite{DDAS} & - & 97.30 $\pm$ .10 & - & 83.40 $\pm$ .2 \\ 
SLACK\cite{Slack} & 96.29 $\pm$ .08 & 97.46 $\pm$ .06 & 79.87 $\pm$ .11 & 84.08 $\pm$ .16 \\
MADAO\cite{hataya2022meta} & 93.3 & 97.3 & 77.6 & - \\
\midrule
 \textbf{\OurMethod} & \textbf{96.54 $\pm$ .16} & \textbf{97.66 $\pm$ .09} & \textbf{80.04 $\pm$ .23} & 84.13 $\pm$ .18 \\
\bottomrule \\
\end{tabular}
 \vspace{-5mm}
\end{table}

\subsubsection{ImageNet-100}
Table \ref{tab:imagenet_results} presents a comparison to previous work on ImageNet-100. All methods were reproduced with the exact training configurations of TA for ResNet-18. Since SLACK officially published four generated augmentation policies, we average the performance of all methods over four random seeds. 
\OurMethod~outperforms the Inception-style baseline of manual choice of augmentations by $\textbf{+1.78\%}$, and surpasses the leading DAS methods compared to by up to $\textbf{+0.94\%}$.
Notably, our approach maintained this high level of accuracy despite a reduction in batch size from $128$ to $64$ images in a batch during the search phase, demonstrating its effectiveness in variance reduction of gradients with respect to the expectation over the augmentation policy, see  \cref{subsec:bilevel pipeline}.

\begin{table}
    \centering
    \vspace{-3mm}
    \caption{Top-1 test accuracy (\%) on ImageNet-100 for ResNet-18. All results are obtained following the exact same training recipe and averaged across 4 random seeds, with the 95\% confidence interval denoted by $\pm$.}
    \label{tab:imagenet_results}
    \begin{tabular}{c@{\hspace{4mm}}c@{\hspace{4mm}}c@{\hspace{4mm}}c@{\hspace{4mm}}c@{\hspace{4mm}}c}
    \toprule
         & Baseline & TA(Wide) & SLACK & \textbf{\OurMethod} \\
         \midrule
        ResNet-18 & 84.84 $\pm$ .43 & 85.68 $\pm$ .54 & 86.19 $\pm$ .81 & \textbf{86.62 $\pm$ .12} \\
    \bottomrule \\
    \end{tabular}
\vspace{-10mm}
\end{table}

\subsubsection{Beyond Natural Images}
\label{subsec:domainnet}
For showcasing the robustness of \OurMethod~ to different domains beyond natural images, we experiment with the DomainNet dataset, comprising six distinct domains. We follow SLACK~\cite{Slack} and compare to DomainBed, the ImageNet version of TA(Wide) and SLACK itself. To this end, we evaluate all of them following the same evaluation setting TA's training recipe and average over four random seeds given the four published augmentation policies of SLACK.
The results in \cref{tab:domainnet_results} show \OurMethod's superior performance compared to existing DAS methods. Notably, all DAS methods outperformed the DomainBed baseline, showcasing the benefits of automatically tailoring designated data augmentation policies for each individual domain. The fact that \OurMethod~yields improved results for all domains on average, and individually for each one, highlights its ability to automatically generate optimized augmentation policies across domains.

\begin{table*}[h!]
\vspace{-3mm}
  \centering
  \caption{Top-1 test accuracy (\%) on DomainNet for ResNet-18. All results are obtained following the exact same training recipe and averaged across 4 random seeds, with the 95\% confidence interval denoted by $\pm$.}
  \label{tab:domainnet_results}
  \begin{tabular}{l@{\hspace{2mm}}|@{\hspace{2mm}}c@{\hspace{4mm}}c@{\hspace{4mm}}c@{\hspace{4mm}}l}
    \toprule
    & DomainBed & TA (Wide) & SLACK &\textbf{\OurMethod}\\
    \midrule
    Real        & 61.91 $\pm$ .27 & 70.27 $\pm$ .27 & 68.94 $\pm$ .56 &\textbf{71.47 $\pm$ .42}\\
    Quickdraw   & 64.55 $\pm$ .16 & 67.01 $\pm$ .27 & 66.39 $\pm$ .35 &\textbf{68.62 $\pm$ .16}\\
    Inforgraph  & 25.64 $\pm$ .27 & 33.73 $\pm$ .41 & 30.92 $\pm$ .61 &\textbf{34.88 $\pm$ .33}\\
    Sketch      & 58.57 $\pm$ .63 & 65.38 $\pm$ .22 & 63.75 $\pm$ .72 &\textbf{66.74 $\pm$ .26}\\
    Painting    & 57.70 $\pm$ .21 & 64.13 $\pm$ .30 & 62.26 $\pm$ .25 &\textbf{64.65 $\pm$ .24}\\
    Clipart     & 64.07 $\pm$ .60 & 69.77 $\pm$ .29 & 70.92 $\pm$ .38 &\textbf{71.26 $\pm$ .26}\\
    \midrule
    Average     & 55.40 $\pm$ .16 & 61.71 $\pm$ .12 & 60.53 $\pm$ .21 &\textbf{62.93 $\pm$ .12}\\
    \bottomrule
  \end{tabular}
\vspace{-5mm}
\end{table*}

\subsection{Ablation Study and Empirical Analysis}
\label{subsec:ablations}

In this section, we analyze the impact of each component of \OurMethod. 
We aim to demonstrate the superiority of jointly learning all four degrees of freedom, and compare the effects of employing a fixed-size augmentation policy against a soft distribution over policy depth. Our analysis also includes a comparison of two augmentation types learning methods, Gumbel-Sinkhorn and Gumbel-Softmax, to illustrate their impact on repetitive sampling of transformations, and later on the final model's performance.

\subsubsection{Joint Learning of All Degrees of Freedom}
\label{sssec:ablation-joint-learning}

We evaluate the benefits of jointly optimizing all four DoFs by freezing each one's distribution as a uniform distribution while learning the other three.
Table~\ref{tab:ablation-joint-learning} shows the results, indicating that the joint learning of all four degrees yields superior results compared to any variant where one is held frozen.
\begin{table}
    \centering
    \caption{The benefit of jointly learning all DoFs. Top-1 test accuracy (\%) on CIFAR100 for WRN-40-2 with different variants of \OurMethod. The results are averaged over 5 random seeds, with the 95\% confidence interval denoted by $\pm$.}
    \label{tab:ablation-joint-learning}
    \begin{tabular}{@{\hspace{1mm}}l@{\hspace{4mm}}|@{\hspace{4mm}}c@{\hspace{4mm}}c@{\hspace{4mm}}c@{\hspace{4mm}}|@{\hspace{4mm}}l@{\hspace{1mm}}}
    \toprule
         \multicolumn{1}{@{\hspace{1mm}}l|@{\hspace{4mm}}}{Frozen degrees of freedom} & $\mu$ & $\Pi$ & $\delta$ & \multicolumn{1}{c}{Top-1} \\
          \midrule
          Frozen uniform magnitudes             & \xmark & \cmark & \cmark & 79.64 $\pm$ .25 \\
          Frozen uniform types \& order         & \cmark & \xmark & \cmark & 79.54 $\pm$ .45 \\
          Frozen uniform depth                  & \cmark & \cmark & \xmark & 79.61 $\pm$ .13 \\
          \midrule
          Joint Learning (\OurMethod)           & \cmark & \cmark & \cmark & \textbf{80.04 $\pm$ .23}\\
    \bottomrule
    \end{tabular}
\end{table}

\subsubsection{Fixed vs. Variable Policy Depth}
\label{sssec:ablation-fixed-vs-soft-depth}

\begin{figure}[tb]
	\centering
	\includegraphics[width=0.6\linewidth]{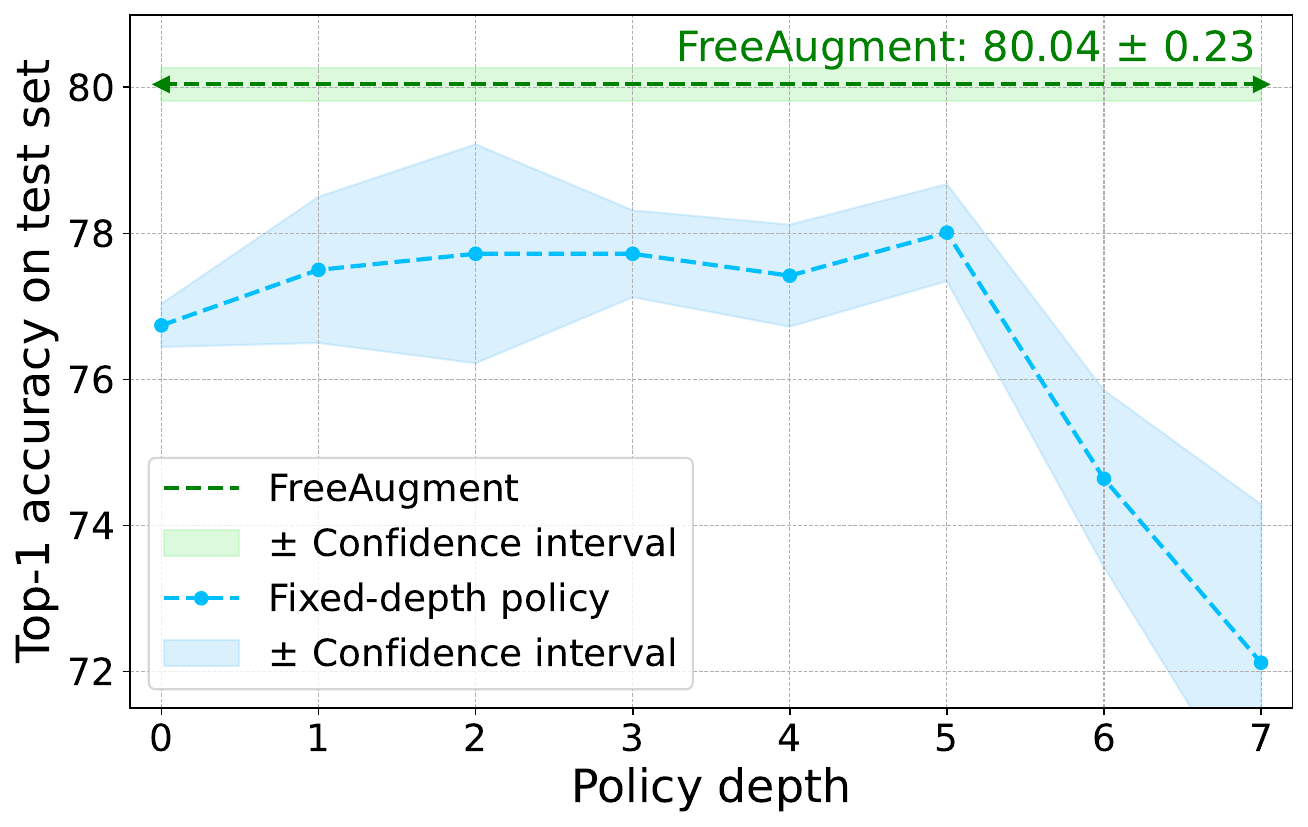}
	\caption{The benefits of a variable policy depth. Top-1 test accuracy on CIFAR100 with WRN-40-2 obtained by \OurMethod~compared to fixed-depth policies. Each point quantifies the average accuracy over 5 random seeds, with the shaded areas indicating the 95\% confidence intervals. The horizontal line marks the performance of \OurMethod~of learned variable policy depth.}
	\label{subfig:ablation-fixed-depth}
 \vspace{-5mm}
\end{figure}

\OurMethod~samples from a learnable probability distribution over the depth of the data augmentation pipeline. 
Figure~\ref{subfig:ablation-fixed-depth} shows the benefit of this approach compared to fixing the depth to each of the possible depths in the search space, while all other DoFs are learned.
The graph shows individual marks indicating the Top-1 Accuracy for each fixed depth on the test set of CIFAR100, and a horizontal line for \OurMethod~of learned variable depth. 
The results are averaged across 5 random seeds, with the 95\% confidence intervals in shadow.
This result highlights the impact of explicitly learning the probability distribution over the number of transformations.

\subsubsection{Mitigating the Repetition of Transformations}
\label{sssec:ablation-gumbel-sinkhorn-vs-softmax}
\begin{figure}[t]
	\centering
	\includegraphics[width=.95\linewidth]{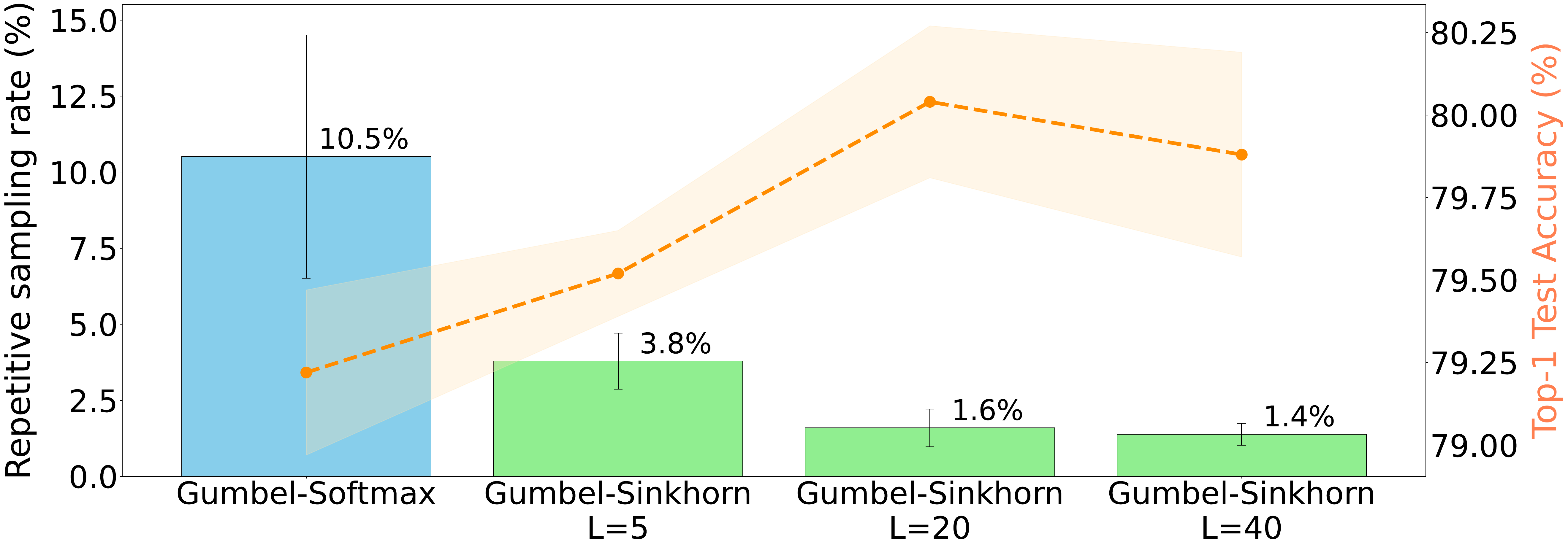}
	\caption{Repetitive sampled transformations rates (Left Y-axis) and  Top-1 test accuracy (\%) on CIFAR100 (Right Y-axis) of the Gumbel-Softmax and Gumbel-Sinkhorn operations with an increasing number of Sinkhorn iterations $L$ for the later. Repetitive transformations rates decreases significantly and accuracy increases in correspondence with the utilization of Gumbel-Sinkhorn with increasing length.}
	\label{subfig:ablation-repetitive-sampling}
\end{figure}

We formulate the choice of types and order of transformations as learning the distribution over permutations via the Gumbel-Sinkhorn operation (Section~\ref{subsec: Augmentations Order}), while other methods~\cite{DifferentiableRandAugment, DeepAA, Slack, DDAS} learn the frequencies of transformations by optimizing categorical distributions separately for every layer in the data augmentation pipeline. 
The later approach allows sampling the same transformation at several layers, while the former mitigates this repetition of transformations, as shown in Figure~\ref{subfig:ablation-repetitive-sampling}.
In this figure, we compare \OurMethod~to its variant, where the Gumbel-Sinkhorn operation is replaced by a series of Gumbel-Softmax operations for sampling from independent categorical distributions across the layers. We measure the rates at which each variant samples repetitive transformations per image along the 10k samples in the evaluation phase of WRN-40-2 on CIFAR10 and the final Top-1 accuracy, also for different amount of Sinkhorn iterations $L$.

The results clearly demonstrate that incorporating Gumbel-Sinkhorn significantly reduces repetitive augmentation sampling by an order of magnitude compared to the variant utilizing Gumbel-Softmax. Increasing the number of Sinkhorn iterations $L$ refines the approximated samples of permutation matrices and thus further reduces the total repetitions of augmentations and its variance, while increasing the accuracy. Saturated encountered at $L=20$.

\section{Conclusion}

In this paper, we propose a novel formulation of the data augmentation search space that allows for the first time to jointly optimize all degrees of freedom of a data augmentation pipeline in an end-to-end manner.
This is enabled by utilizing fully differentiable tools for learning the probability distributions over discrete options for the number of transformations in the pipeline, their continuous magnitudes, and their discrete types and order. We jointly learn (1) the categorical distribution over the first by utilizing the Gumbel-Softmax reparameterization trick, (2) the parameterized distribution over magnitudes by applying differentiable implementations of the transformation, and (3) the distribution over their order and types using the Gumbel-Sinkhorn operator. All are learned in an end-to-end manner by alternating gradient decent steps to solve a bilevel optimization problem, whose objective is that the augmentations make a model trained on the train set generalize better to a held-out validation set. Extensive experiments show that this effective way of optimizing the joint probability distribution over all degrees of freedom yields data augmentation policies that achieve state-of-the-art results across various benchmarks.
\newpage

%
%
\bibliographystyle{splncs04}
\bibliography{references}

\newpage

\appendix

\section*{Supplementary Material}
\label{sec:supplamentry}

    \section{Policy Search Cost}
\cref{tab:search-cost} summarizes the policy search cost for different DAS methods. While achieving better or comparable performance with all of those, \OurMethod~finds augmentation policies with a relatively small cost, especially considering that the cost is reported for a single search, while some other methods have to repeatedly apply it for several policy depth values. Some other methods are limited to searching merely two consecutive transformations due to the exponential growth of the cost with the policy depth. \OurMethod~is the only one to eliminate that entirely, requiring a single search to obtain all degrees of freedom, including policy depth, without resulting in iterative solutions.

\begin{table*}[h!]
\centering
\caption{Policy search cost in GPU hours on CIFAR10/100 with WRN40-2. Each method is reported with its corresponding hardware and time. -: missing information. \\
\textsuperscript{*}Reported in DDAS.
\textsuperscript{**}Exponential complexity in policy depth and hence limited to 2 transformations only.
\textsuperscript{***}Requires additional search on policy depth.}
\label{tab:search-cost}
    \begin{tabular}{@{\hspace{2mm}}l@{\hspace{4mm}}c@{\hspace{4mm}}c@{\hspace{4mm}}l@{\hspace{2mm}}}
    \toprule
        \multicolumn{1}{@{\hspace{2mm}}l}{Method} & 
        \multicolumn{1}{c@{\hspace{2mm}}}{GPU hours} & 
        \multicolumn{1}{c@{\hspace{2mm}}}{Hardware} & Architecture\\
        \midrule
        AA\cite{AutoAugment}~\textsuperscript{**} & 5000 & Tesla P100 & WRN-40-2 \\
        PBA\cite{PBA}~\textsuperscript{**} & 5 & Titan XP & WRN-40-2 \\
        FastAA\cite{FastAA}~\textsuperscript{**} & 3.5 & Tesla V100  & WRN-40-2 \\
        DADA\cite{DADA}~\textsuperscript{**} & 0.1/0.2 & Titan XP  & WRN-28-10 \\
        RA\cite{RandAugment}~\textsuperscript{*} \textsuperscript{**} & 33 & RTX 2080Ti  & WRN-28-10 \\
        FasterAA\cite{FasterAA}~\textsuperscript{**} & 0.23 & Tesla V100  & WRN-40-2 \\
        DeepAA\cite{DeepAA} & 9 & -  & WRN-28-10 \\
        DRA\cite{DifferentiableRandAugment}~\textsuperscript{***} & 0.4 & Tesla P100  & WRN-28-10 \\
        DDAS\cite{DDAS}~\textsuperscript{***} & 0.15 & RTX 2080Ti  & WRN-28-10 \\
        SLACK\cite{Slack}~\textsuperscript{***} & 4 & -  & WRN-40-2 \\ 
        MADAO\cite{hataya2022meta}~\textsuperscript{***} & 1.7 & -  & WRN-28-2 \\ 
        \OurMethod & 1.2 & RTX A6000  & WRN-40-2 \\
    \bottomrule
    \end{tabular}
\end{table*}

\newpage
\section{Found Augmentation Policies}
\begin{figure}[h!]
	\centering
	\includegraphics[width=.7\linewidth]{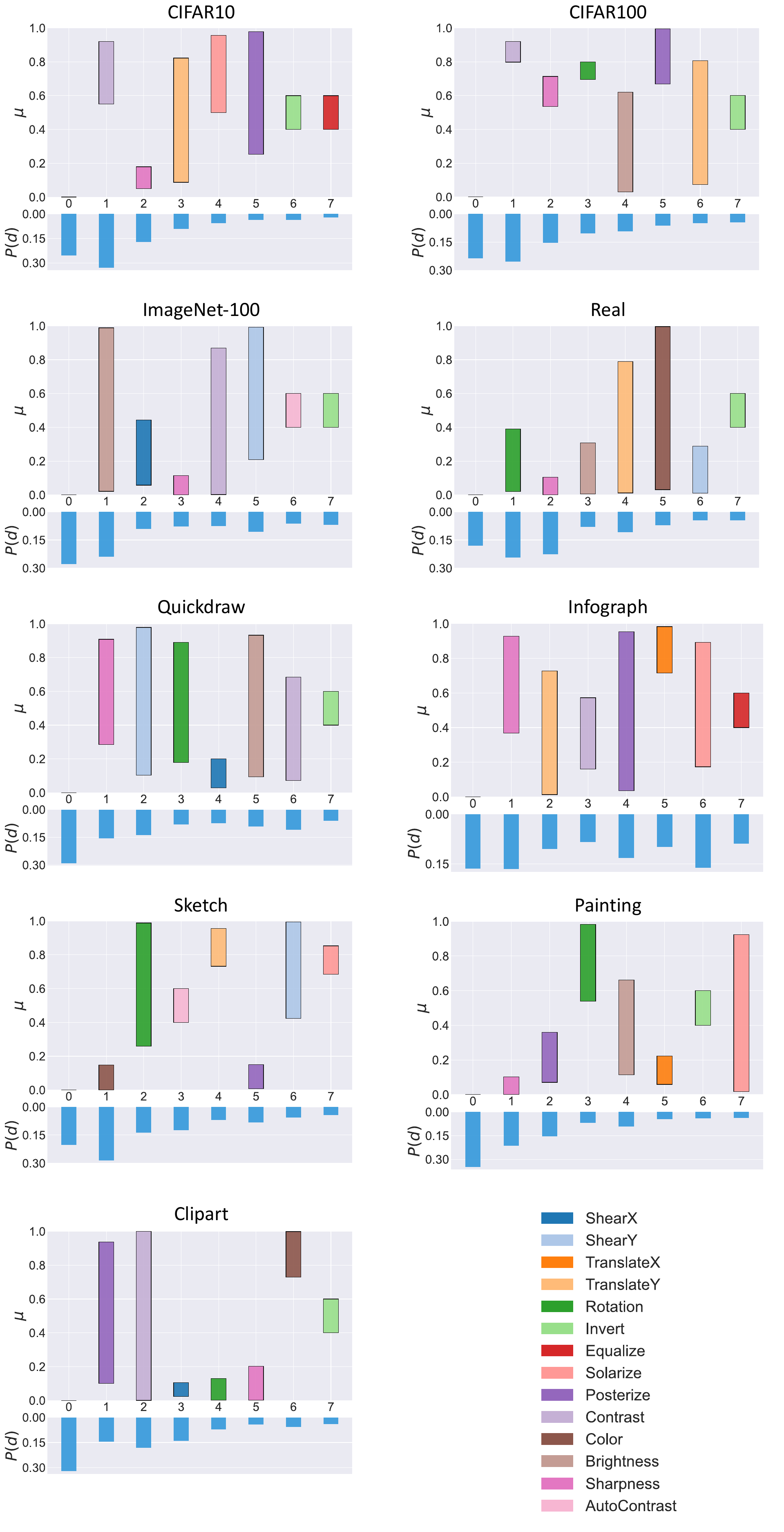}
	\caption{Found augmentation policies on CIFAR10/100, ImageNet-100, and DomainNet. Transformations that are not associated with a learnable magnitude are depicted with a magnitude range of $[0.4,0.6]$}
	\label{subfig:found-policies}
 \vspace{-10mm}
\end{figure}

\cref{subfig:found-policies} visualizes the data augmentation policies found by \OurMethod.

 \vspace{-3mm}

\begin{figure}[p]
    \centering
    \begin{subfigure}[b]{0.45\linewidth}
        \includegraphics[width=\linewidth]{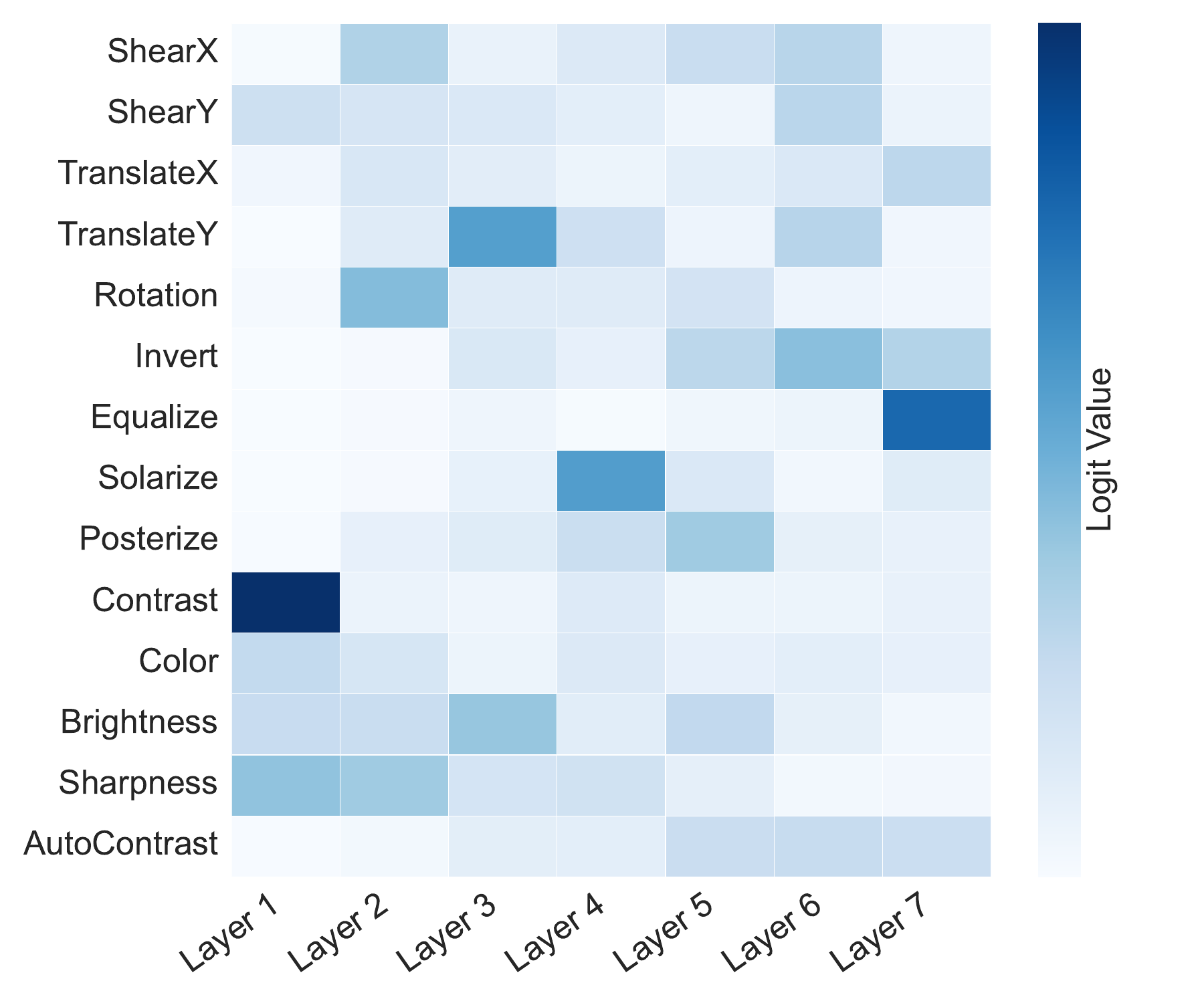}
        \caption{Found policy distribution over transformation types and order}
    \end{subfigure}
    \hfill
    \begin{subfigure}[b]{0.45\linewidth}
        \includegraphics[width=\linewidth]{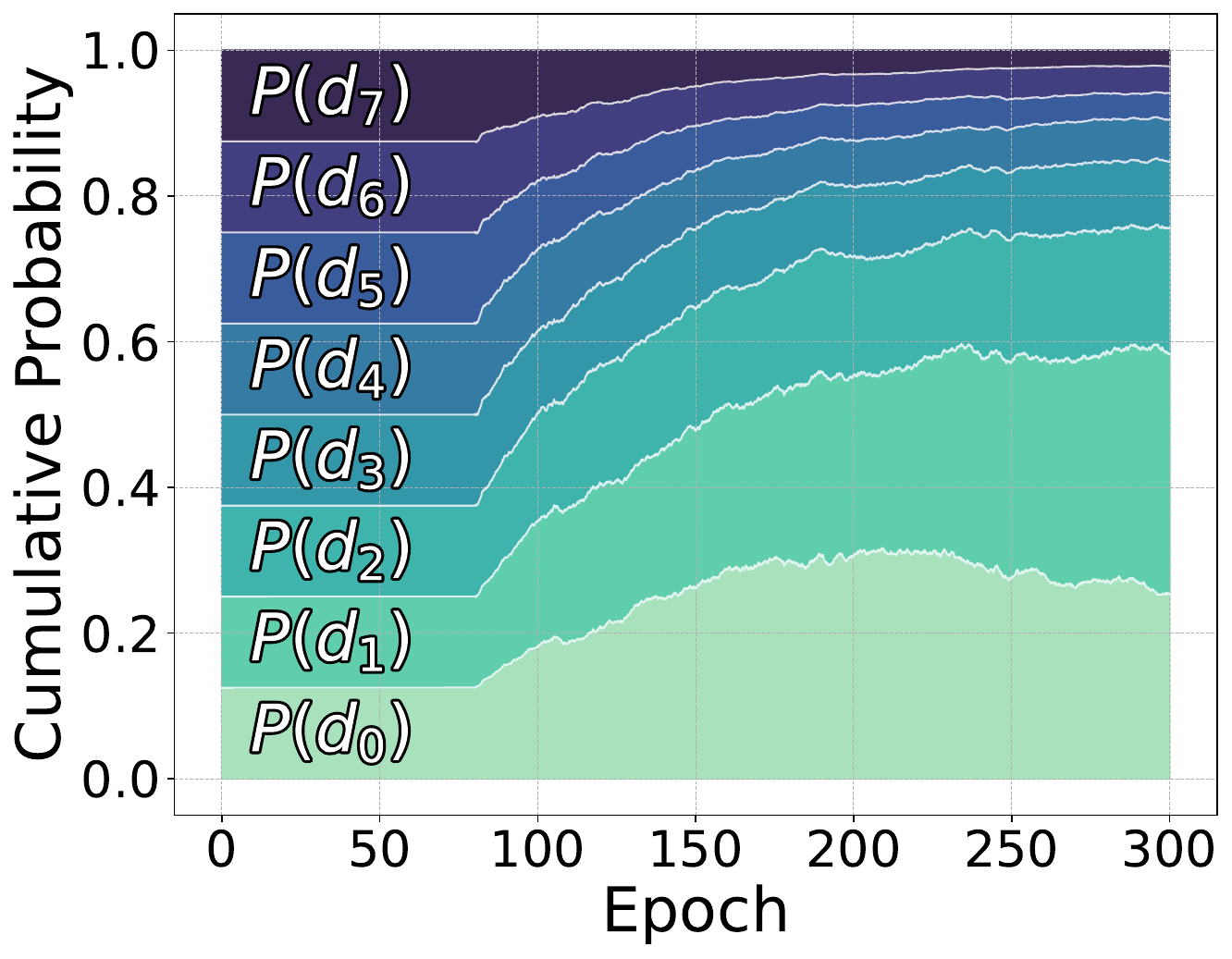}
        \caption{Policy depth dynamics}
    \end{subfigure}

    \vspace{0.5cm}
    
    \begin{subfigure}[b]{0.32\linewidth}
        \includegraphics[width=\linewidth]{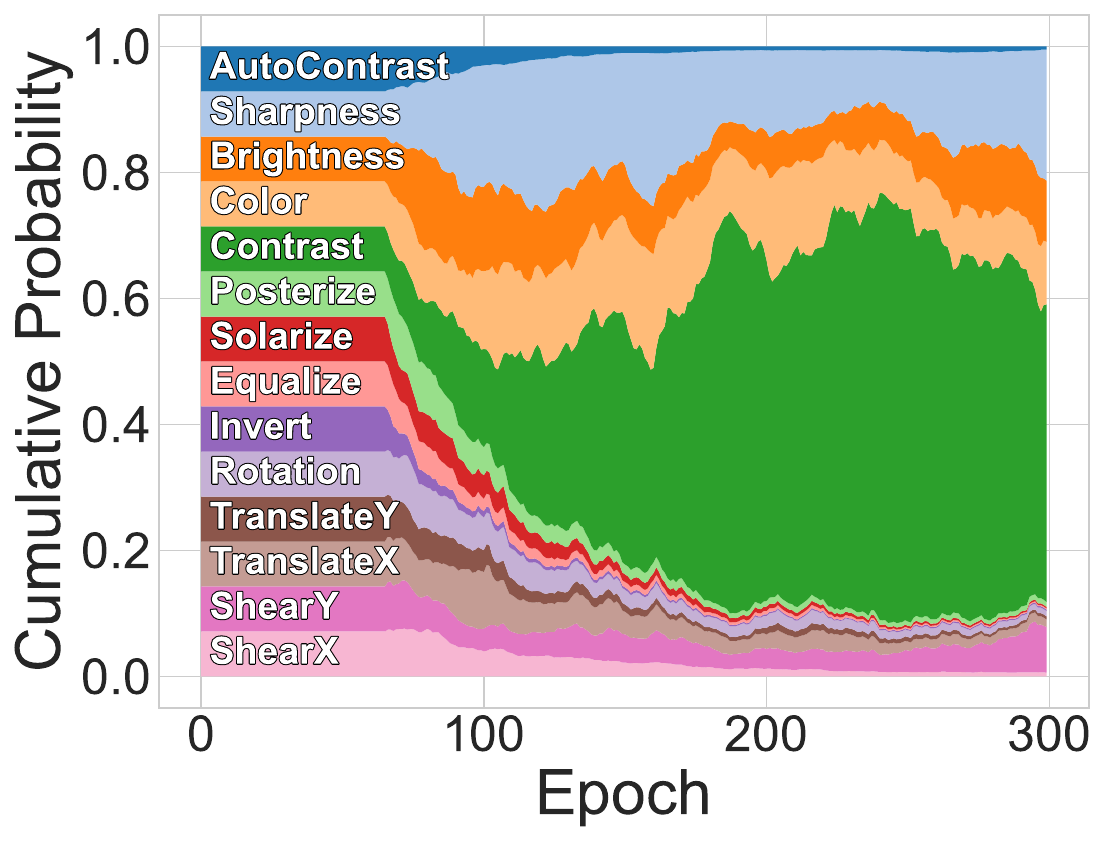}
        \caption{Layer 1}
    \end{subfigure}
    \hfill
    \begin{subfigure}[b]{0.32\linewidth}
        \includegraphics[width=\linewidth]{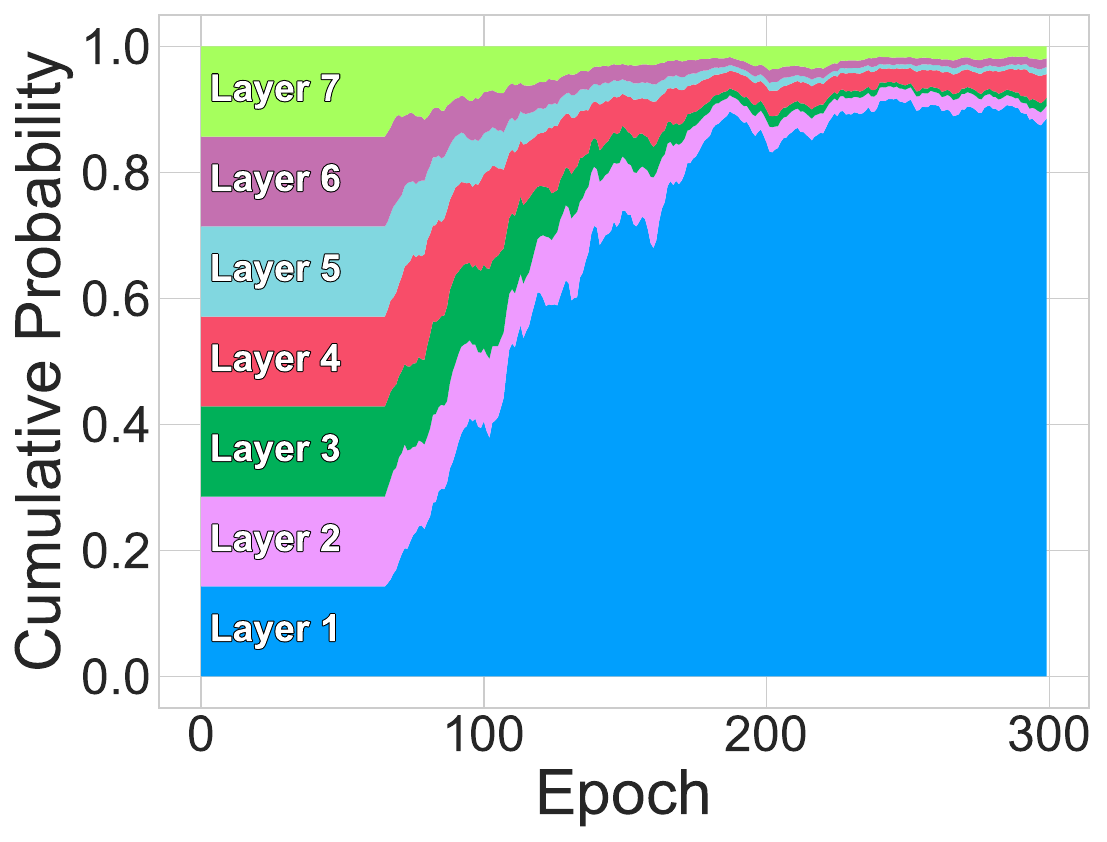}
        \caption{Contrast across layers}
    \end{subfigure}
    \hfill
    \begin{subfigure}[b]{0.32\linewidth}
        \includegraphics[width=\linewidth]{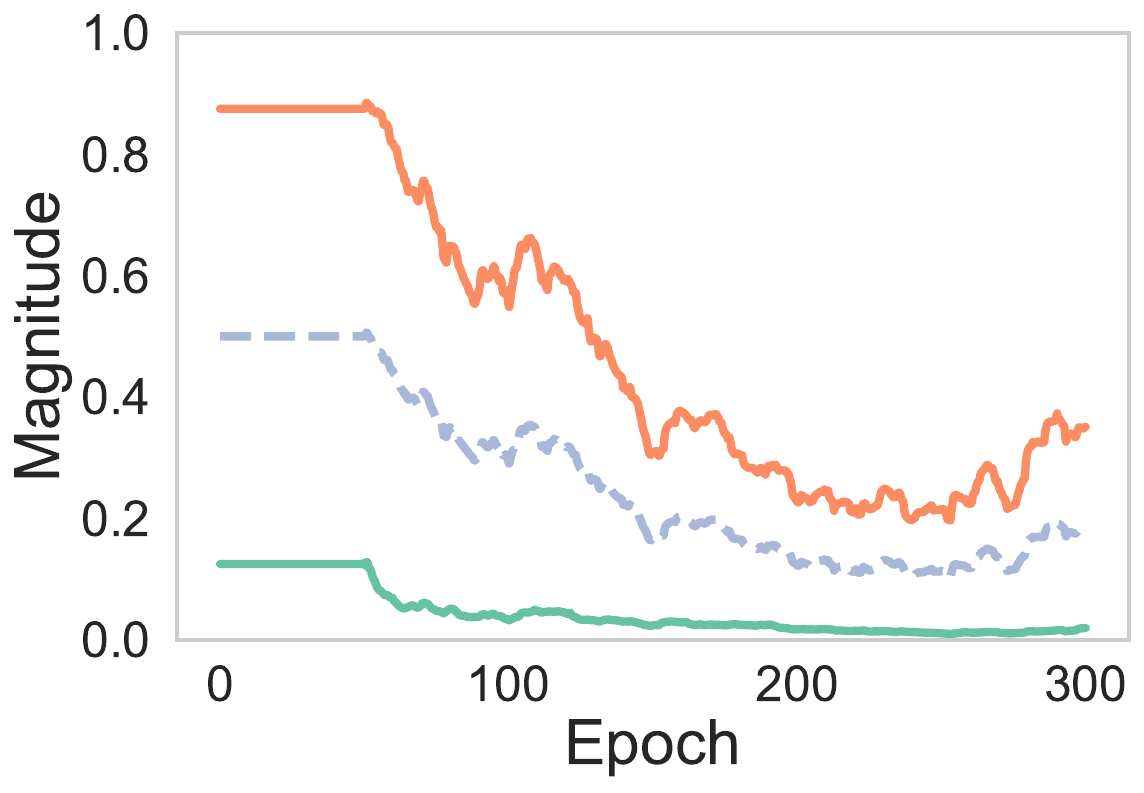}
        \caption{Contrast magnitude}
    \end{subfigure}
    
    \vspace{0.5cm}
    
    \begin{subfigure}[b]{0.32\linewidth}
        \includegraphics[width=\linewidth]{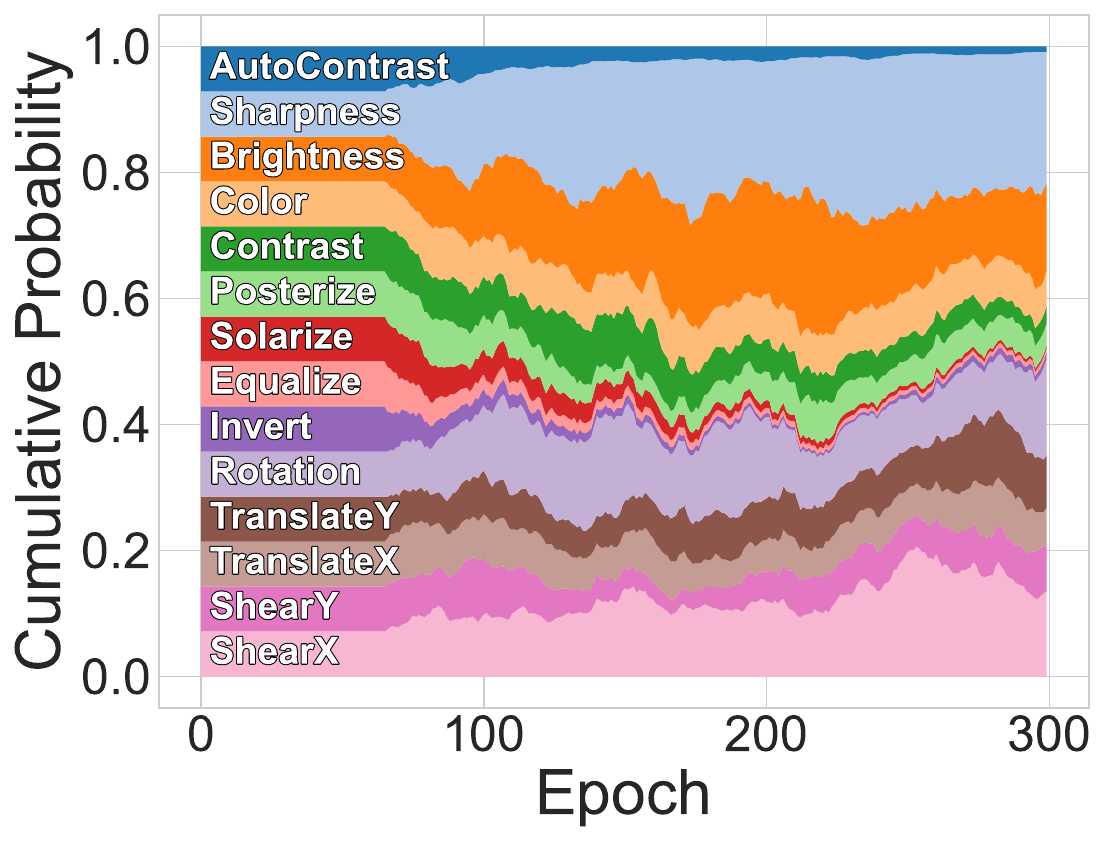}
        \caption{Layer 2}
    \end{subfigure}
    \hfill
    \begin{subfigure}[b]{0.32\linewidth}
        \includegraphics[width=\linewidth]{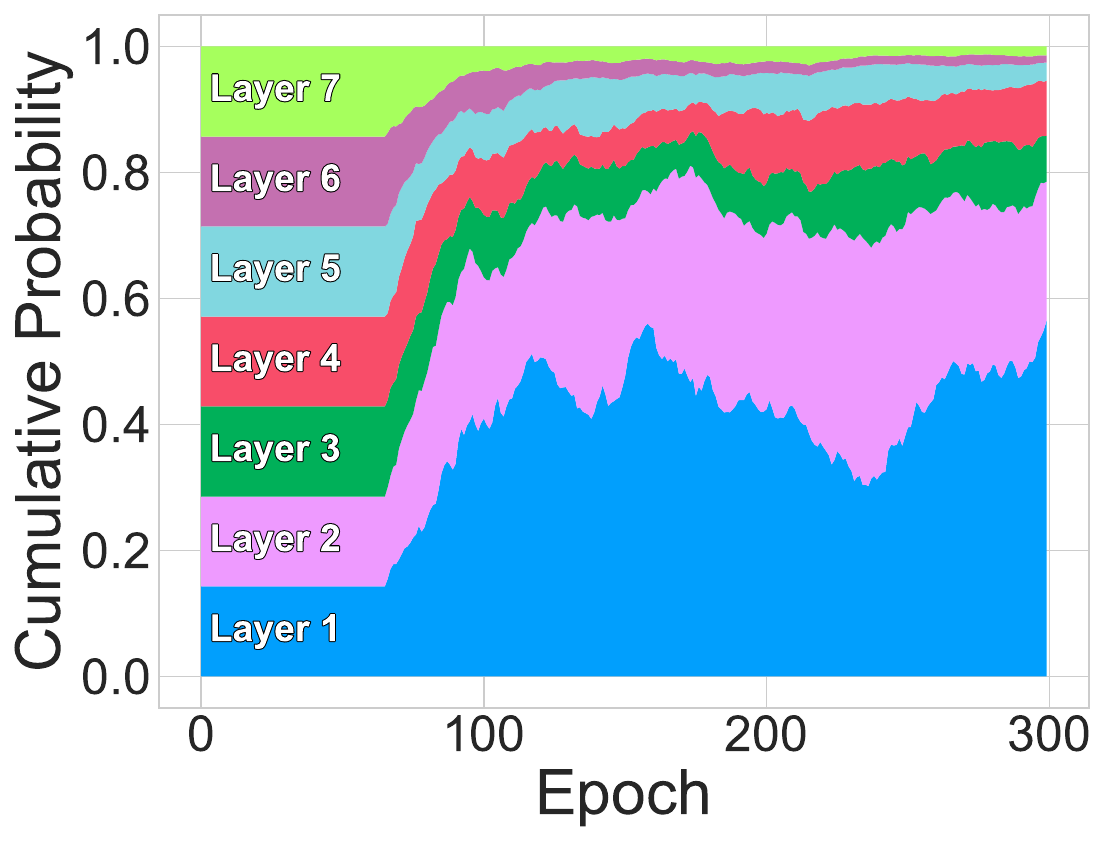}
        \caption{Sharpness across layers}
    \end{subfigure}
    \hfill
    \begin{subfigure}[b]{0.32\linewidth}
        \includegraphics[width=\linewidth]{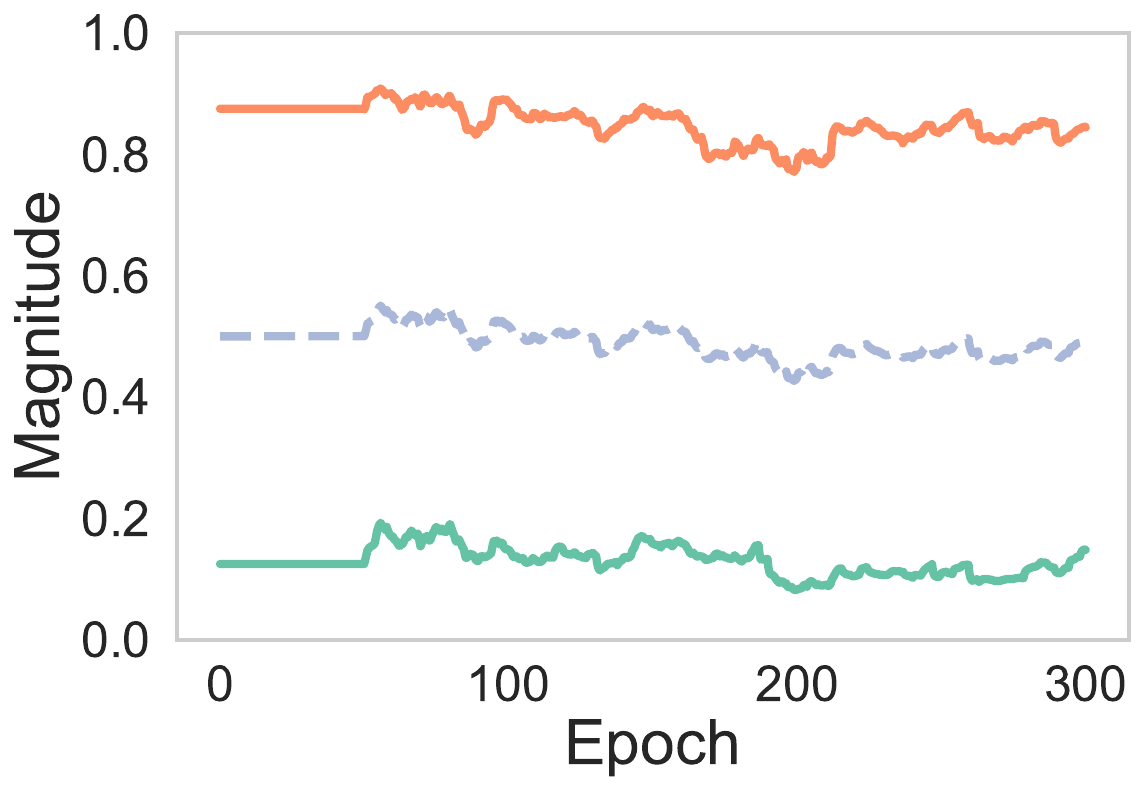}
        \caption{Sharpness magnitude}
    \end{subfigure}
    
    \vspace{0.5cm}
    
    \begin{subfigure}[b]{0.32\linewidth}
        \includegraphics[width=\linewidth]{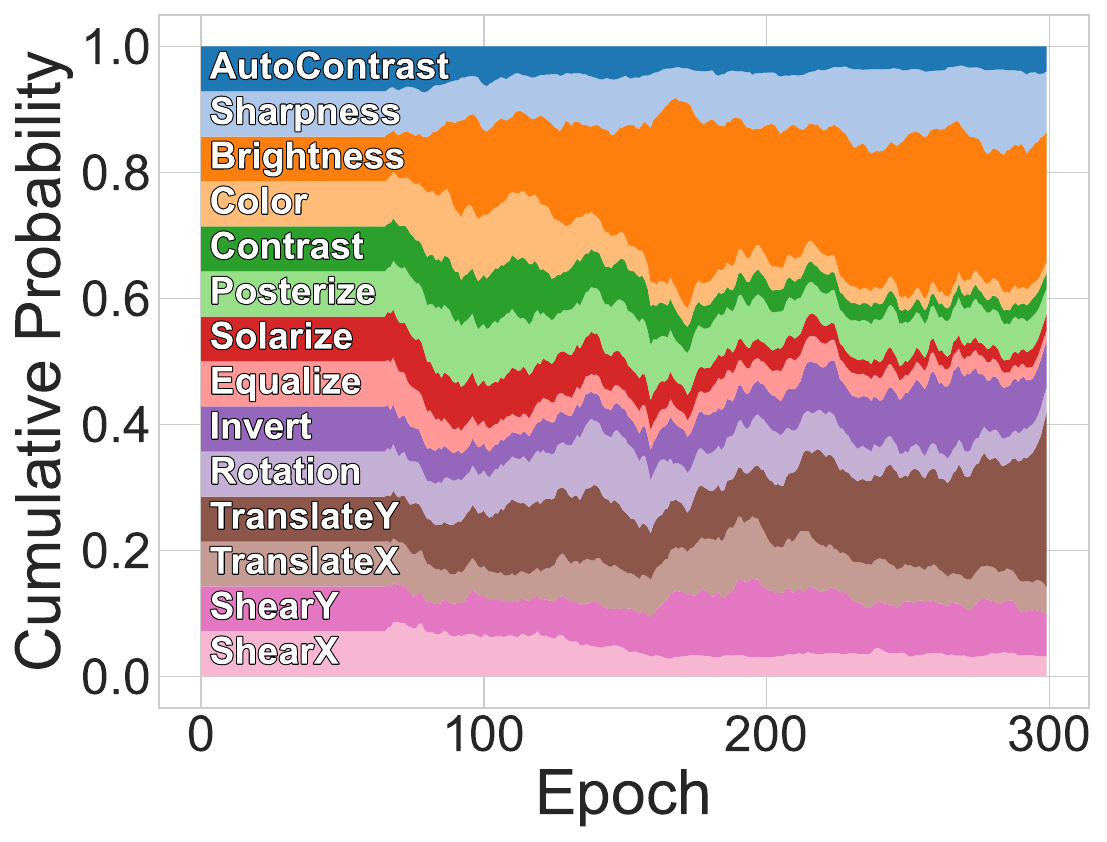}
        \caption{Layer 3}
    \end{subfigure}
    \hfill
    \begin{subfigure}[b]{0.32\linewidth}
        \includegraphics[width=\linewidth]{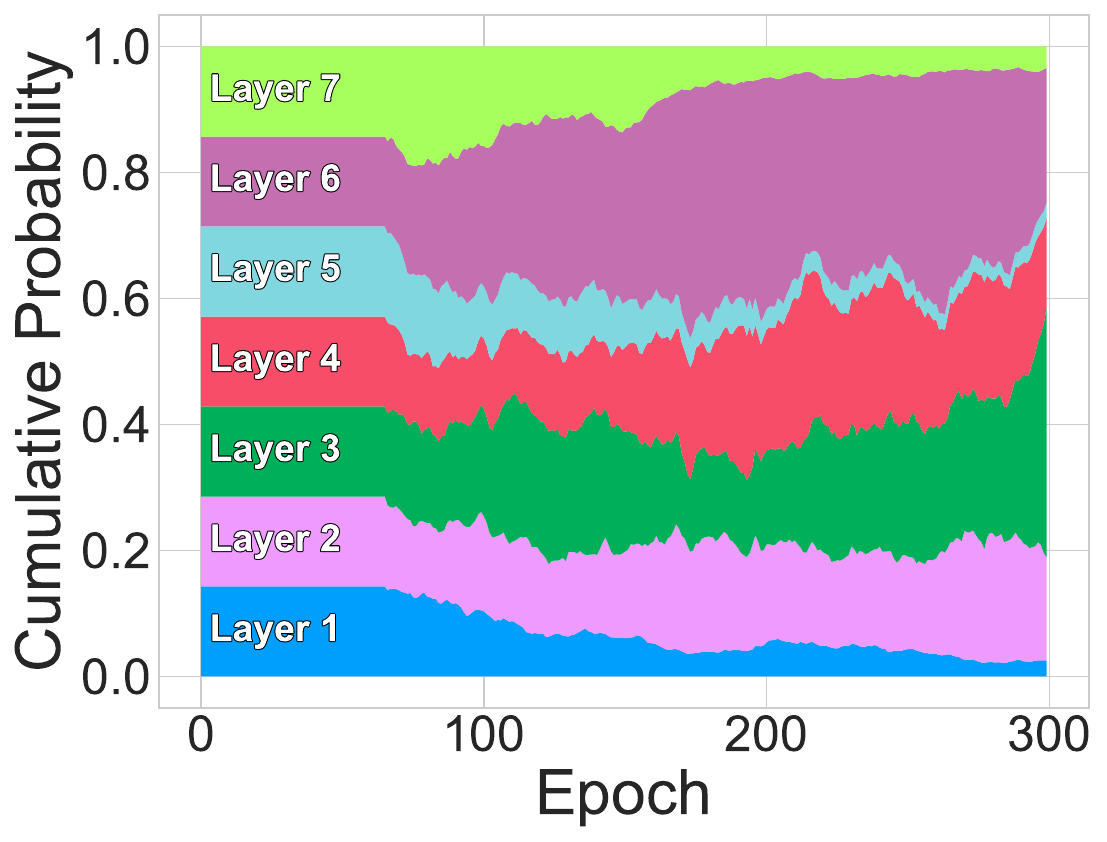}
        \caption{TranslateY across layers}
    \end{subfigure}
    \hfill
    \begin{subfigure}[b]{0.32\linewidth}
        \includegraphics[width=\linewidth]{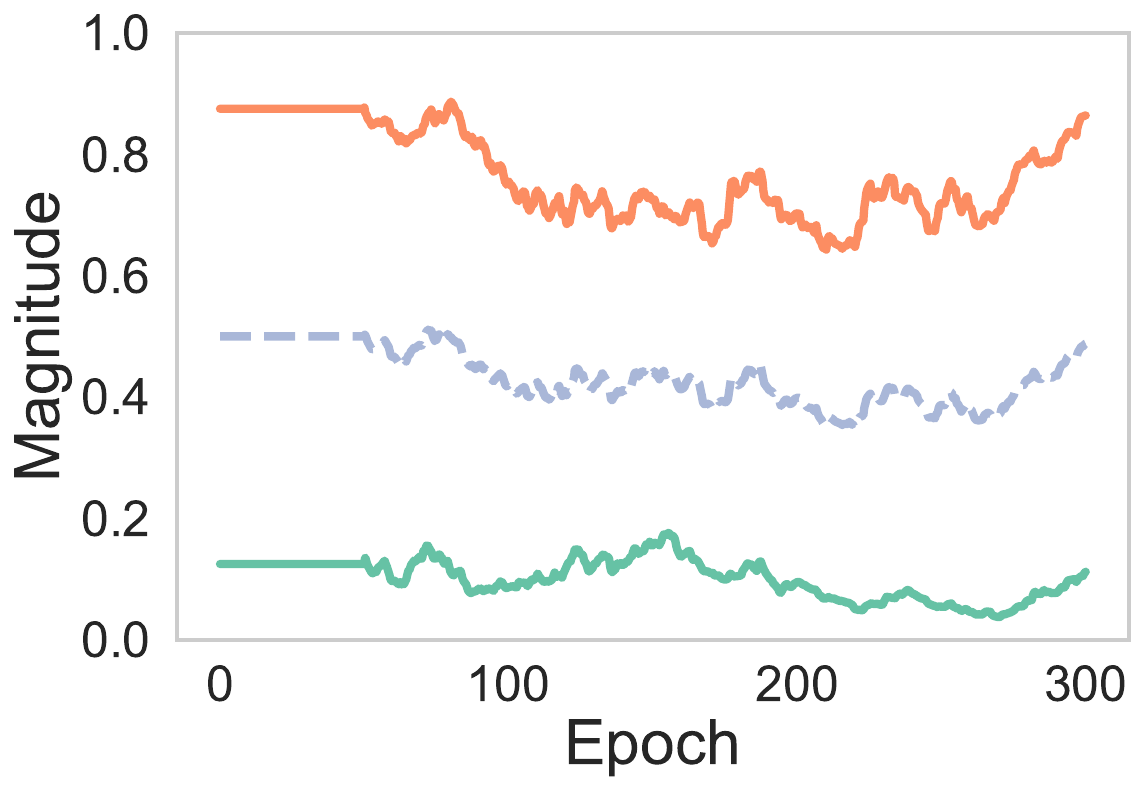}
        \caption{TranslateY magnitude}
    \end{subfigure}
    
    \caption{Policy dynamics during the search phase of CIFAR10 (Part 1/2)}
\end{figure}

\begin{figure}[p]
    \centering
    \begin{subfigure}[b]{0.32\linewidth}
        \includegraphics[width=\linewidth]{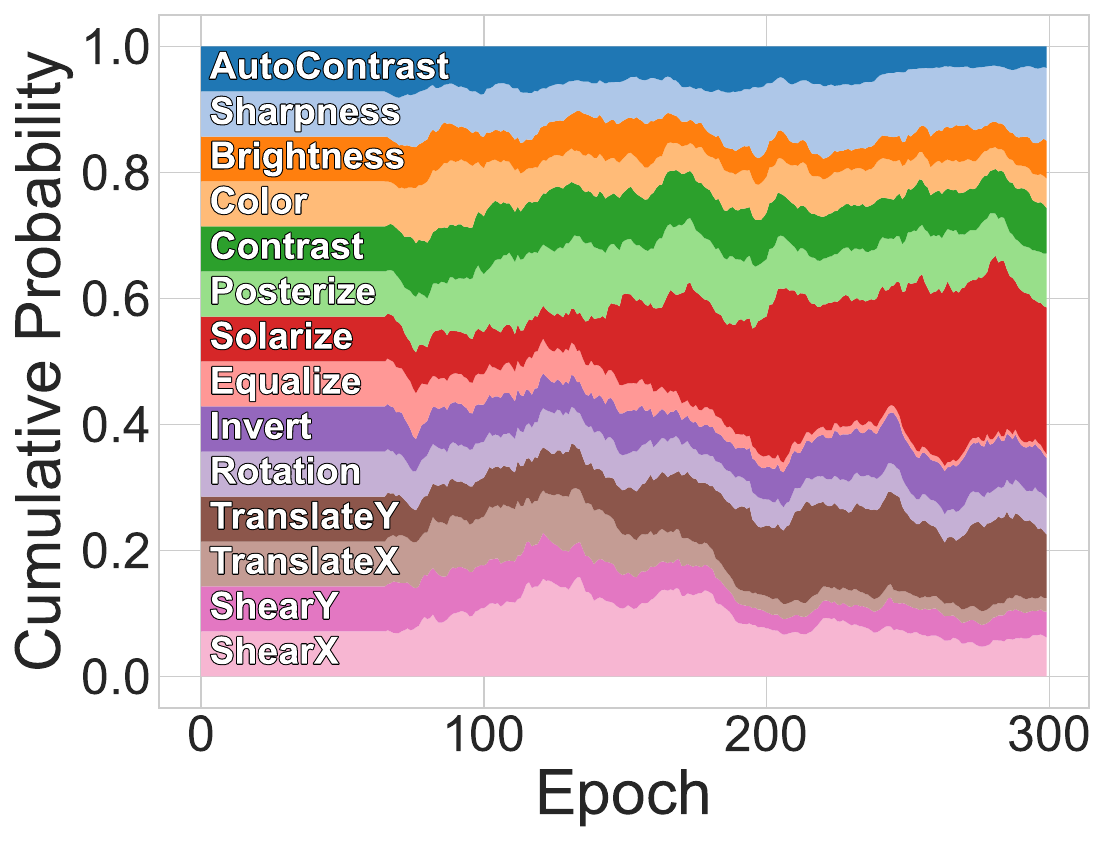}
        \caption{Layer 4}
    \end{subfigure}
    \hfill
    \begin{subfigure}[b]{0.32\linewidth}
        \includegraphics[width=\linewidth]{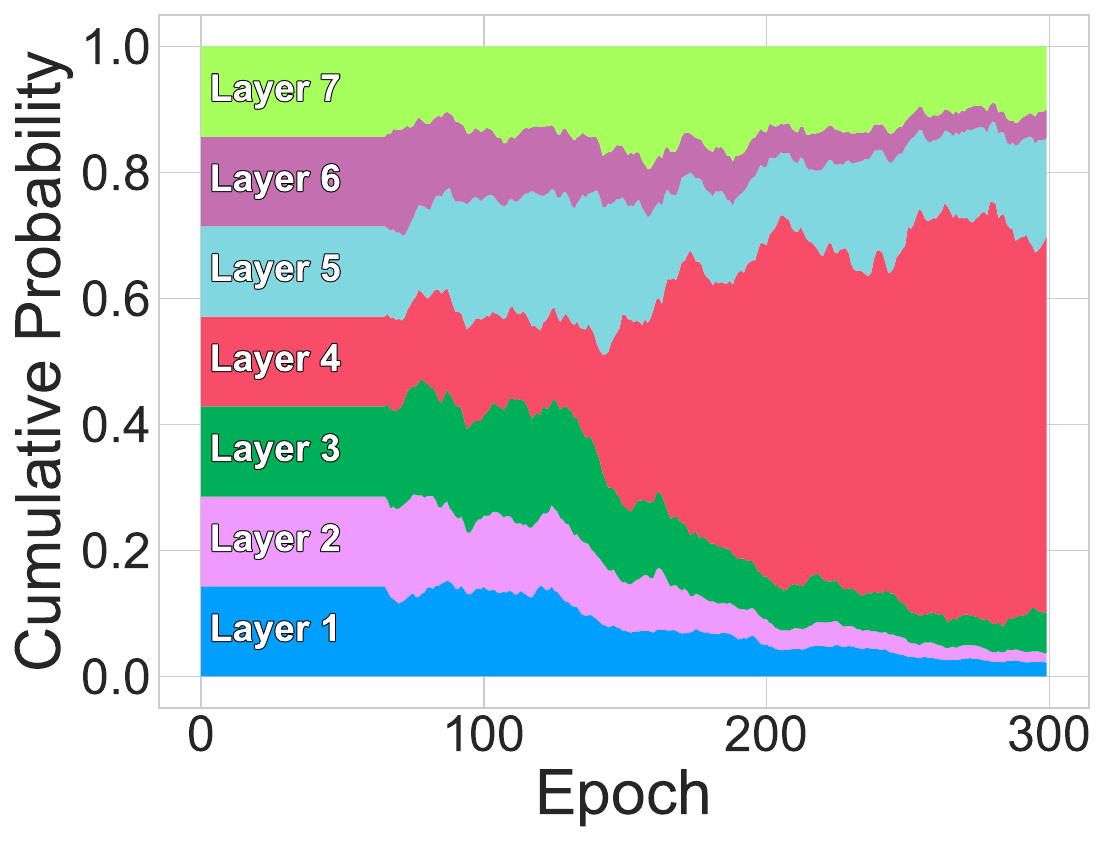}
        \caption{Solarize across layers}
    \end{subfigure}
    \hfill
    \begin{subfigure}[b]{0.32\linewidth}
        \includegraphics[width=\linewidth]{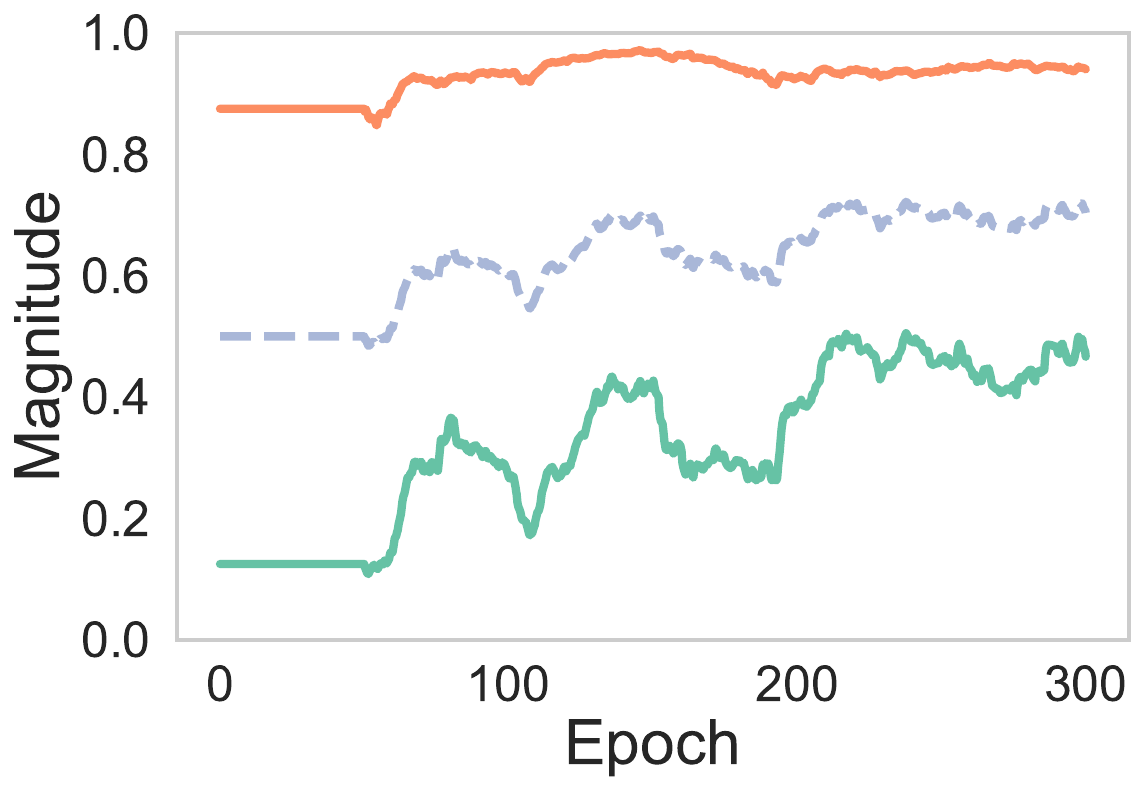}
        \caption{Solarize magnitude}
    \end{subfigure}

    \vspace{0.5cm}
    
    \begin{subfigure}[b]{0.32\linewidth}
        \includegraphics[width=\linewidth]{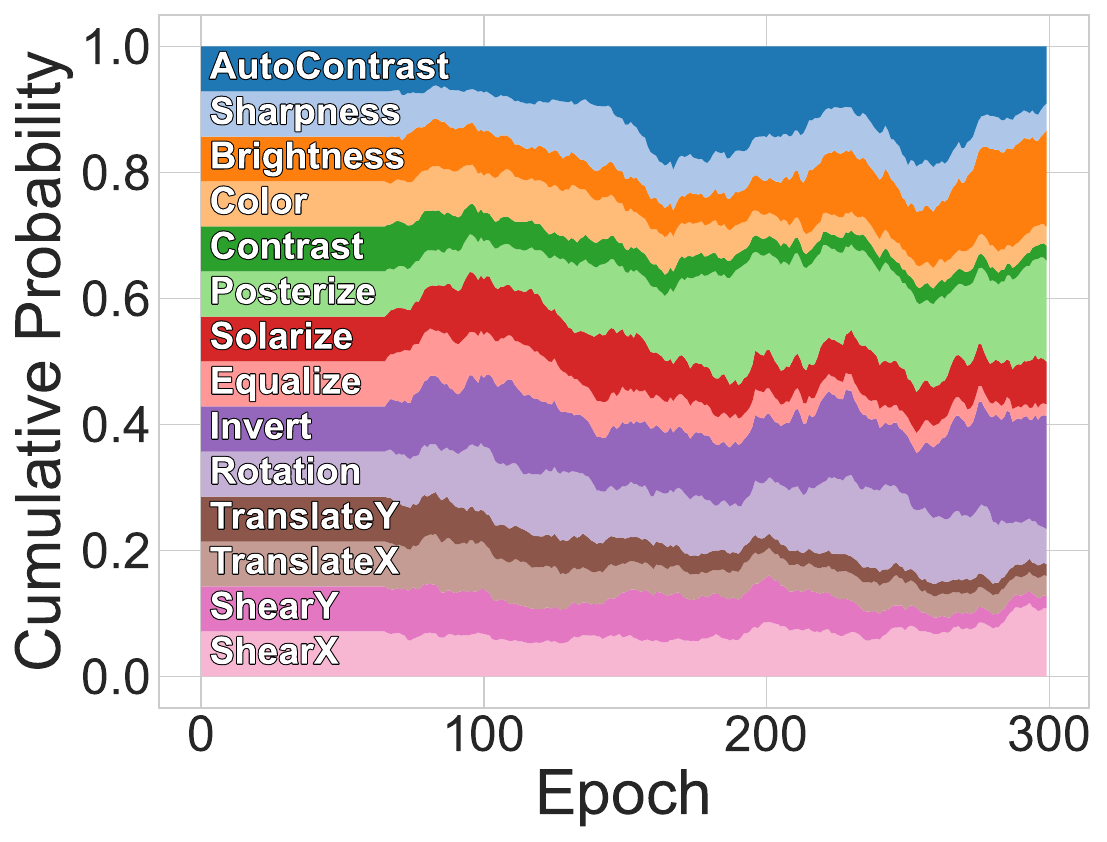}
        \caption{Layer 5}
    \end{subfigure}
    \hfill
    \begin{subfigure}[b]{0.32\linewidth}
        \includegraphics[width=\linewidth]{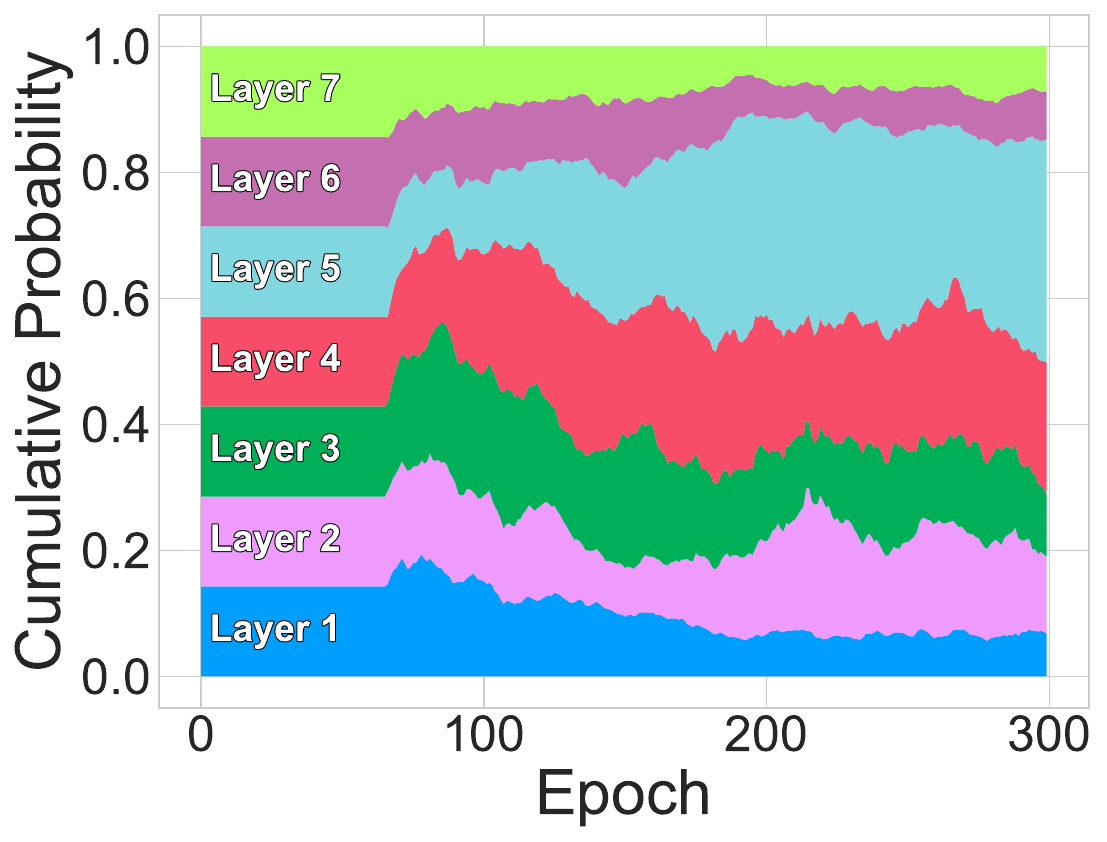}
        \caption{Posterize across layers}
    \end{subfigure}
    \hfill
    \begin{subfigure}[b]{0.32\linewidth}
        \includegraphics[width=\linewidth]{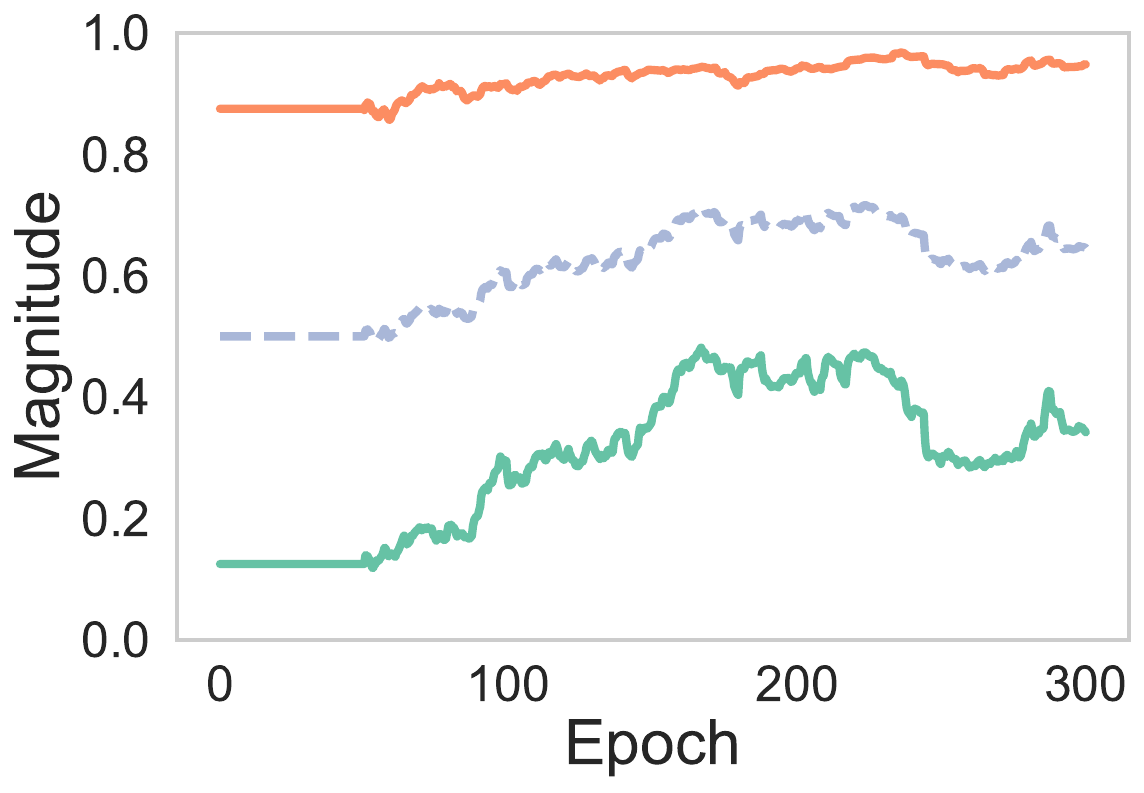}
        \caption{Posterize magnitude}
    \end{subfigure}
    
    \vspace{0.5cm}
    
    \begin{subfigure}[b]{0.32\linewidth}
        \includegraphics[width=\linewidth]{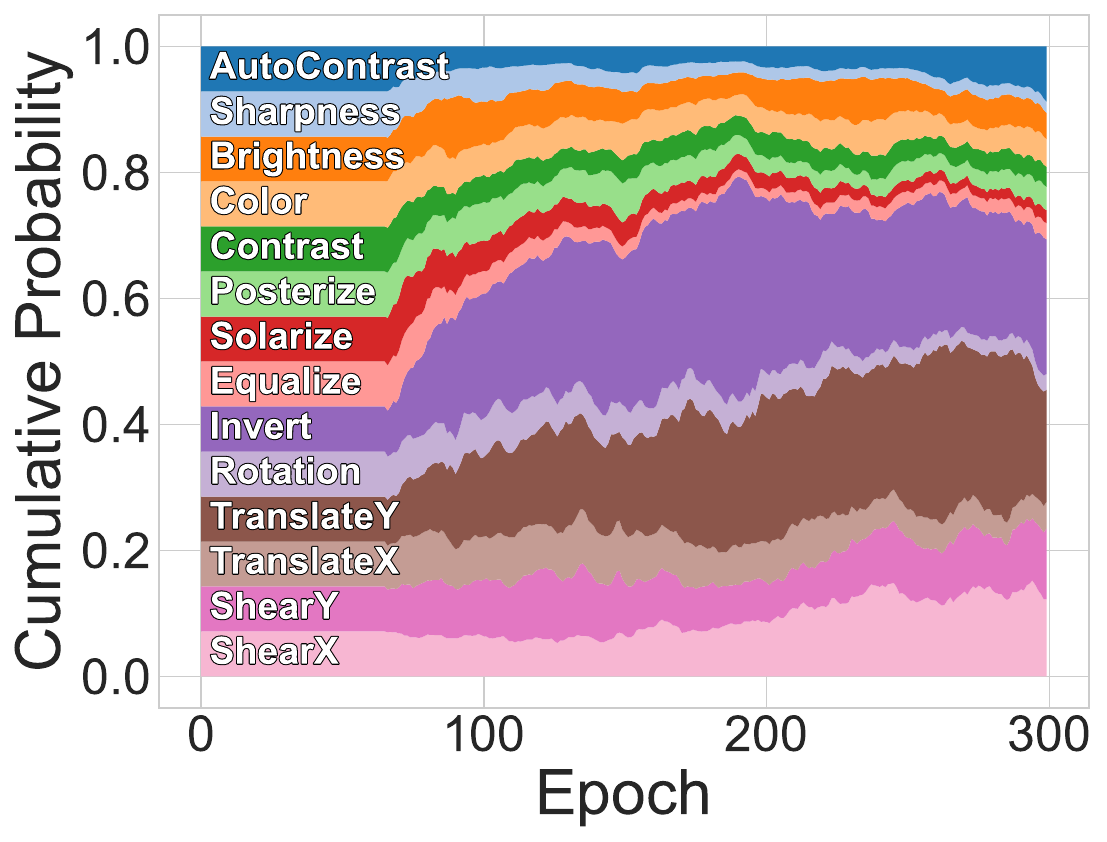}
        \caption{Layer 6}
    \end{subfigure}
    \hfill
    \begin{subfigure}[b]{0.32\linewidth}
        \includegraphics[width=\linewidth]{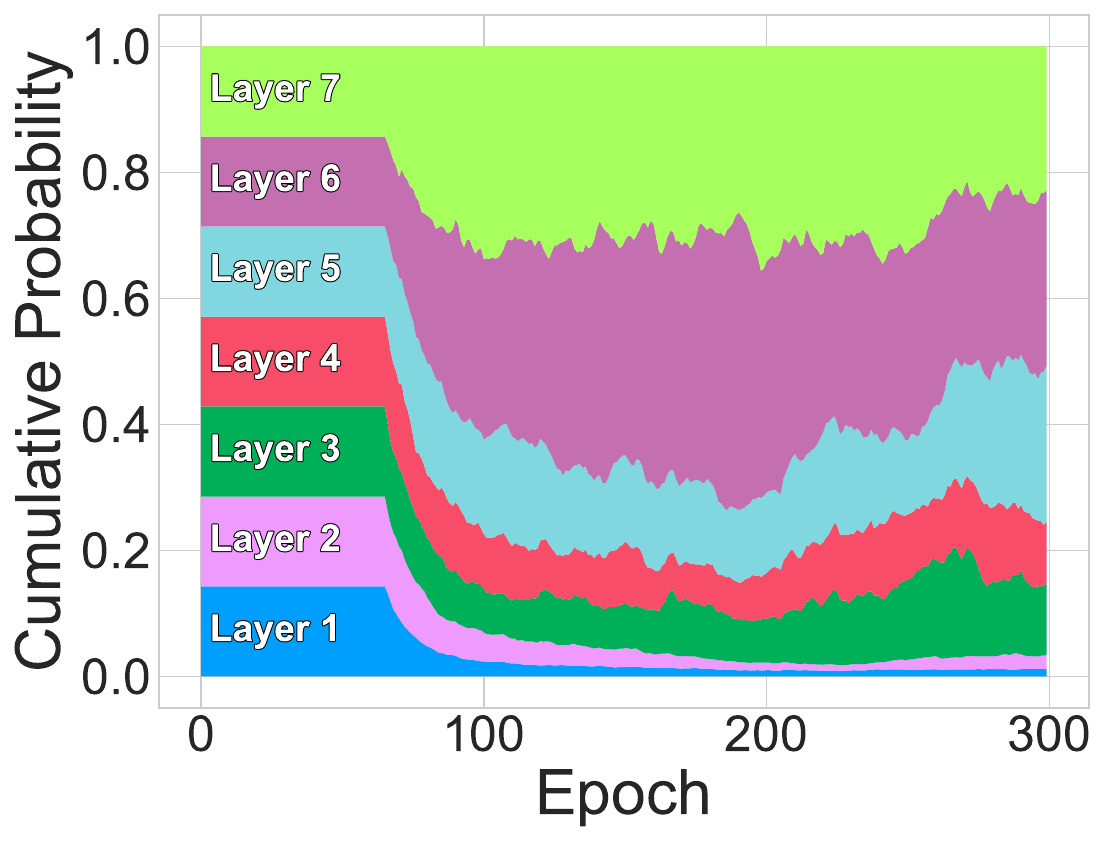}
        \caption{Invert across layers}
    \end{subfigure}
    \hfill
    \begin{subfigure}[b]{0.32\linewidth}
        \includegraphics[width=\linewidth]{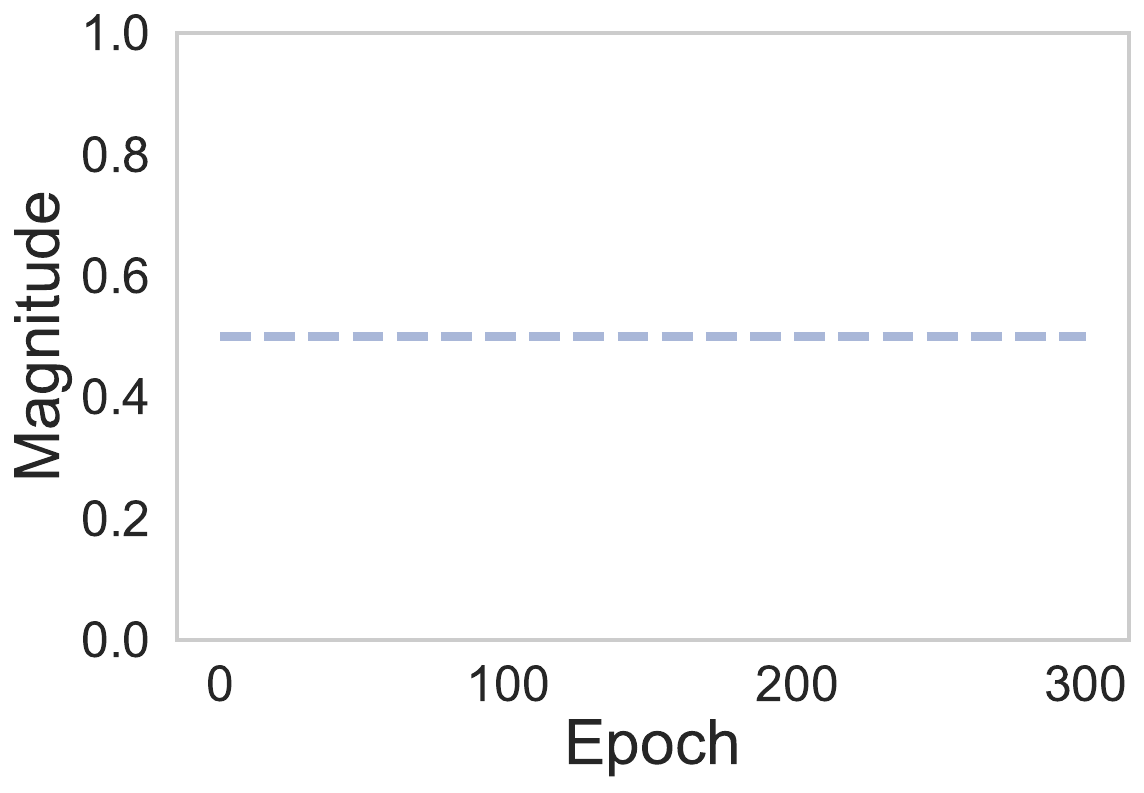}
        \caption{Invert}
    \end{subfigure}
    
    \vspace{0.5cm}
    
    \begin{subfigure}[b]{0.32\linewidth}
        \includegraphics[width=\linewidth]{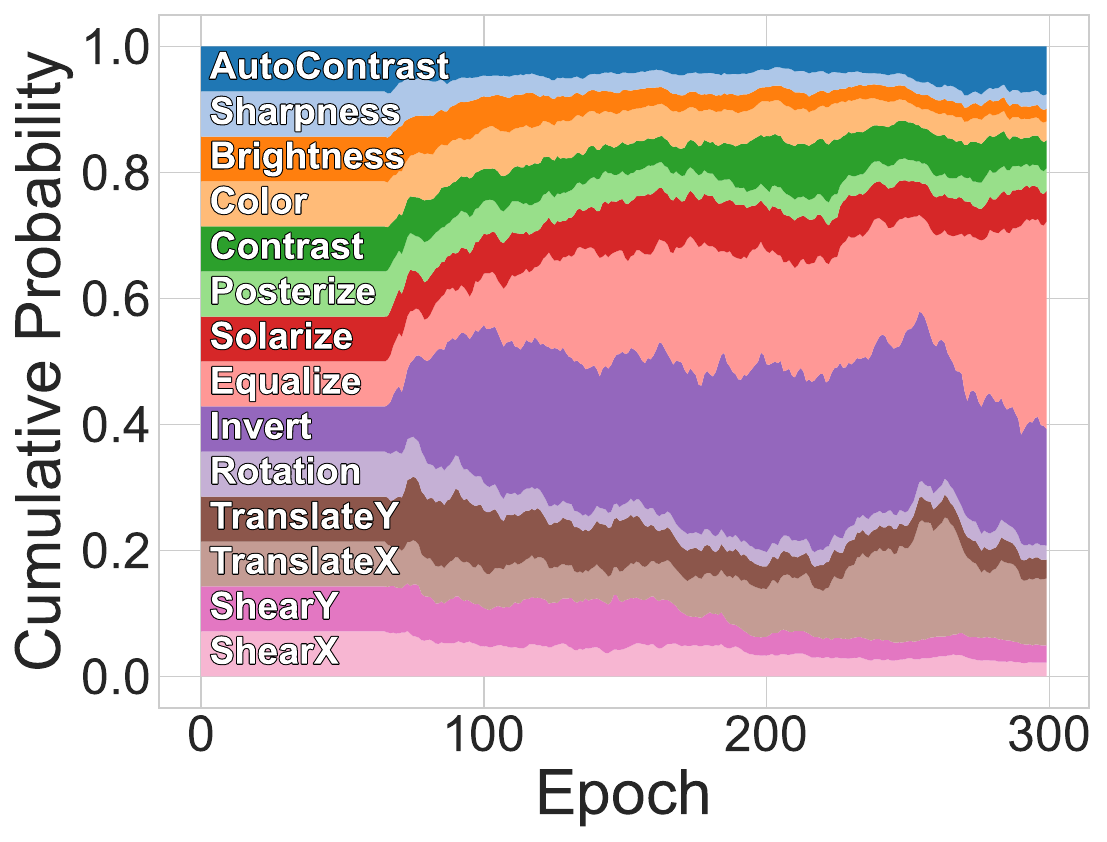}
        \caption{Layer 7}
    \end{subfigure}
    \hfill
    \begin{subfigure}[b]{0.32\linewidth}
        \includegraphics[width=\linewidth]{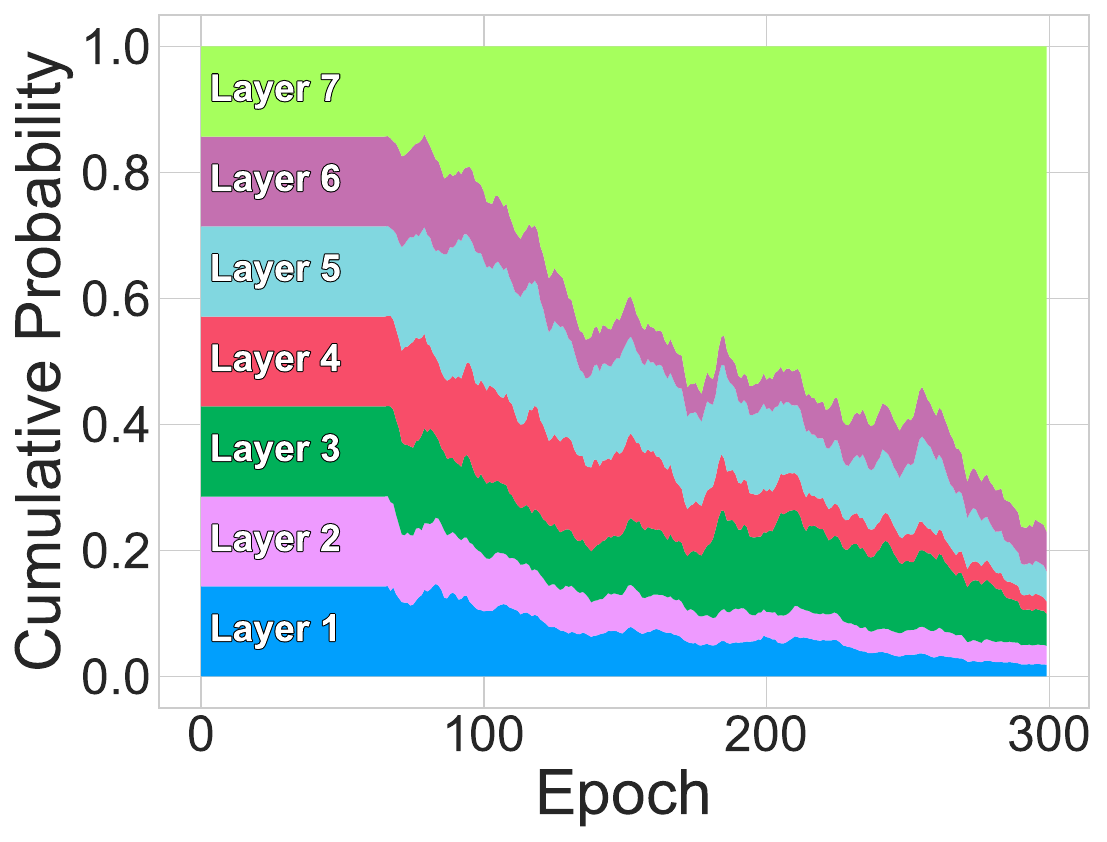}
        \caption{Equalize across layers}
    \end{subfigure}
    \hfill
    \begin{subfigure}[b]{0.32\linewidth}
        \includegraphics[width=\linewidth]{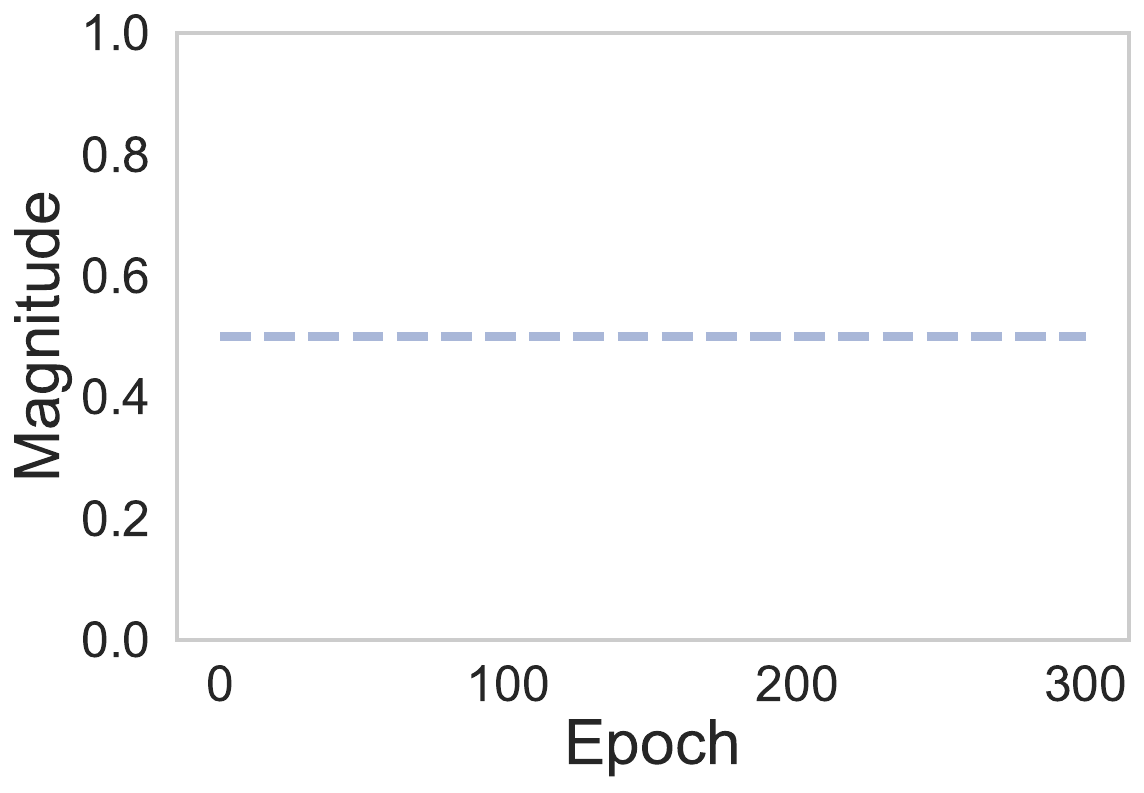}
        \caption{Equalize}
    \end{subfigure}
    
    \caption{Policy dynamics during the search phase of CIFAR10 (Part 2/2)}
\end{figure}

\section{Hyper-Parameters and Training Configurations}
\label{supplementary:training-configurations}
The configurations for policy search and evaluation of each dataset can be found in \cref{tab:search-hyper-parameters} and \cref{tab:evaluation-hyper-parameters}, respectively. Following previous work, both in search and evaluation, the classifier's learning rate is decayed according to a cosine annealing scheduler. 
Following TA\cite{TrivialAugment}, for CIFAR10/100 images we apply pad-and-crop to a resolution of $32 \times 32$ with a reflection mode and a pad length of 4, random horizontal flipping with a probability of 0.5, then our generated policy is applied, followed by cutout \cite{devries2017improved} with a length of 16. For ImageNet-100 and DomainNet we use random resized crops and scales between 0.08 and 1.0 to a resolution of $224 \times 244$ using bicubic interpolation, random horizontal flipping with a probability of 0.5, then our generated policy is applied, followed by cutout with a length of 75, which is about 1/3 of the image size, as done in \cite{TIMM}.

\begin{table}[htb]
    \centering
    \scriptsize
    \caption{Policy search hyper-parameters. Classifier hyper-parameters are identical to those in the policy evaluation.}
    \label{tab:search-hyper-parameters}
    \begin{tabular}{c@{\hspace{4mm}}c@{\hspace{4mm}}c@{\hspace{4mm}}c}
    \toprule
        & CIFAR10/100 & ImageNet-100 & DomainNet \\
        \midrule
        Architecture & WRN-40-2, WRN-28-10 & ResNet-18 & ResNet-18 \\
        Epochs & 300 & 300 & 300 \\
        Batch size & 128 & 64 & 64 \\
        \midrule
        Classifier Hyper-Parameters & & $-$ Same as in \cref{tab:evaluation-hyper-parameters} $-$ & \\
        \midrule
        Policy optimizer & Adam & Adam & Adam \\
        $\mu$ learning rate & 0.02& 0.02& 0.02\\
        $\Pi$ learning rate & 0.01& 0.01& 0.01\\
        $\delta$ learning rate & 1.0& 1.0& 1.0\\
    \bottomrule
    \end{tabular}
    \vspace{-5mm}
\end{table}

\begin{table}[htb]
    \centering
    \scriptsize
    \caption{Policy evaluation hyper-parameters.}
    \label{tab:evaluation-hyper-parameters}
    \begin{tabular}{c@{\hspace{4mm}}c@{\hspace{4mm}}c@{\hspace{4mm}}c}
    \toprule
        & CIFAR10/100 & ImageNet-100 & DomainNet \\
        \midrule
        Architecture & WRN-40-2, WRN-28-10 & ResNet-18 & ResNet-18 \\
        Epochs & 200 & 270 & 200 \\
        Batch size & 128 & 256 & 128 \\
        \midrule
        Optimizer & SGD & SGD & SGD \\
        Learning rate & 0.1 & 0.1 & 0.1 \\
        Weight decay & 0.0005 & 0.0001 & 0.0001 \\
        Momentum parameter & 0.9 & 0.9 & 0.9 \\
        Nesterov Momentum & True & True & True \\
    \bottomrule
    \end{tabular}
    \vspace{-5mm}
\end{table}

\newpage
\section{Search Space Details}
\label{supplementary:search-space}
\cref{tab:magnitude-ranges} lists the transformations in the search space of \OurMethod~with their corresponding magnitude ranges. All magnitude parameters correspond to Kornia's image transformations\footnote{\href{https://kornia.readthedocs.io}{https://kornia.readthedocs.io}}.
\begin{table*}[h!]
\centering
\caption{List of transformations in the search space and their corresponding magnitude range. Note that AutoContrast, Invert, and Equalize do not have magnitude parameters.}

\label{tab:magnitude-ranges}
    \begin{tabular}{@{\hspace{2mm}}l@{\hspace{4mm}}|@{\hspace{4mm}}c@{\hspace{2mm}}}
    \toprule
        \multicolumn{1}{@{\hspace{2mm}}l}{Transformation} & 
        \multicolumn{1}{c@{\hspace{2mm}}}{Magnitude Range} \\ 
        \midrule
        ShearX            & [-0.6, 0.6] \\
        ShearY            & [-0.6, 0.6] \\
        TranslateX        & [-0.5, 0.5] \\
        TranslateY        & [-0.5, 0.5] \\
        Rotate            & [-30, 30] \\
        Solarize          & [0.6, 1] \\
        Posterize         & [2, 8] \\
        Contrast          & [0.4, 2.0] \\
        Color             & [0.0, 1.0] \\
        Brightness        & [-0.4, 0.4] \\
        Sharpness         & [0.0, 2.0] \\
        AutoContrast      & - \\
        Invert            & -\\
        Equalize          & -  \\
    \bottomrule
    \end{tabular}
\end{table*}

\section{Additional Ablations}

\subsection{Sample single augmentation per-batch vs. per-image}
\OurMethod~samples distinct augmentation sequences for every image in the batch rather than sampling once for the entire batch. \cref{code:per-image-sampling-code} outlines a simple and efficient PyTorch per-image sampling implementation of Algorithm 1 in the main text. 
\cref{tab:ablation-per-image-sampling} compares the performance of \OurMethod~when it samples the same augmentation policy for the entire batch versus per-image. As can be inferred from the table, the per-image sampling helps to reduce the variance of the gradient estimates with respect to $\phi$ during the search and thus leads to better results.

\begin{table}
    \centering
    \caption{Top-1 test accuracy (\%) on CIFAR100 for WRN-40-2 with different variants of \OurMethod's, one that samples different augmentations per-image in the batch, and one that samples the same augmentations for the whole batch. The results are averaged over 5 random seeds, with the 95\% confidence interval denoted by $\pm$.}
    \label{tab:ablation-per-image-sampling}
    \begin{tabular}{l@{\hspace{2mm}}|@{\hspace{2mm}}c}
    \toprule
        \OurMethod~variant & Top-1 Accuracy \\
        \midrule
        Sample augmentation per-batch & 79.43 $\pm$ .23 \\
        Sample augmentation per-image & \textbf{80.04 $\pm$ .23}\\
    \bottomrule
    \end{tabular}
\end{table}

\subsection{No Warm-up For $\phi$}
\cref{tab:ablation-no-warmup} compares the results of a search phase that begins with a warm-up for the learnable policy parameters $\phi$ and a search phase that starts learning those at the very beginning. At the former, $\phi$ are not updated until the classifier is sufficiently learned so that the gradients of $\phi$ backpropagating through it are meaningful. Such strategy aims to avoid noise in the learning process of $\phi$ caused by a premature classifier, yielding a bad approximation of the lower problem of the bilevel optimization. As can be seen from the results, warm-up improves the performance.
\begin{table}
    \centering
    \caption{Top-1 test accuracy (\%) on CIFAR100 for WRN-40-2 where \OurMethod learns with and without warm-up. The results are averaged over 5 random seeds, with the 95\% confidence interval denoted by $\pm$.}
    \label{tab:ablation-no-warmup}
    \begin{tabular}{l@{\hspace{2mm}}|@{\hspace{2mm}}c}
    \toprule
        \OurMethod~variant & Top-1 Accuracy \\
        \midrule
        Without warm-up & 79.53 $\pm$ .13 \\
        With warm-up & \textbf{80.04 $\pm$ .23}\\
    \bottomrule
    \end{tabular}
\end{table}

\subsection{Magnitude Distribution}
\cref{tab:ablation-normal-magnitude} compares the results of \OurMethod~variant that learns the a uniform distribution over each magnitude $m$, versus a Gaussian distribution, such that,
$$m=\hat{\sigma}\cdot \epsilon + \hat{\mu}\quad ; \quad \epsilon \sim \mathcal{N}(0, 1)$$
with $\hat{\mu}, \hat{\sigma}$ being the learnable magnitude parameters and $\mathcal{N}(0, 1)$ is the normal distribution.

The uniform distribution is initialized to $75\%$ of the normalized range as $(0.125,0.875)$, and the normal distribution is initialized with a mean $\hat{\mu}$ of $0.5$ and a standard deviation $\hat{\sigma}$ of $0.1875$. This results in samples falling in the interval of $(0.125,0.875)$ with a probability of $0.95$. As can be seen from the table, the results using both sampling methods are fairly close, with slightly better results using uniform magnitude sampling.

\begin{table}
    \centering
    \caption{Top-1 test accuracy (\%) on CIFAR100 for WRN-40-2 with different variants of \OurMethod's, one that learns a uniform magnitude distribution, and another that learns a normal magnitude distribution. The results are averaged over 5 random seeds, with the 95\% confidence interval denoted by $\pm$.}
    \label{tab:ablation-normal-magnitude}
    \begin{tabular}{l@{\hspace{2mm}}|@{\hspace{2mm}}c}
    \toprule
        \OurMethod~variant & Top-1 Accuracy \\
        \midrule
        Gaussian Magnitudes & 79.9 $\pm$ 0.19 \\
        Uniform Magnitudes & \textbf{80.04 $\pm$ .23}\\
    \bottomrule
    \end{tabular}
\end{table}

\subsection{Depth Search by Probability of Application}

Additional depth-searching strategy treats augmentation application as sampling from a learnable Bernoulli distribution, similar to the methods in \cite{FastAA, FasterAA, DifferentiableRandAugment, MedPipe}
\cref{tab:ablation-depth-search-strategies} compares the results of \OurMethod~variants using different depth-searching strategies.
Learnable application probabilities for each augmentation were initialized to 0.75.
As shown in the table, \OurMethod~with a learnable Gumbel-Softmax distribution over policy depths achieves better performance.
Compared to FreeAugment with a single representation for each policy depth $d$ out of $D$ transformations, learning the application probability for each transformation effectively yields ${D \choose d}$ representations for the same policy depth $d$. The resulting redundancy makes the DAS operate within a much larger representation of the search space without effectively adding more policies.

\begin{table}
    \centering
    \caption{Comparing strategies of depth search. Top-1 test accuracy (\%) on CIFAR100 for WRN-40-2. The results are averaged over 5 random seeds, with the 95\% confidence interval denoted by $\pm$.}
    \label{tab:ablation-depth-search-strategies}
    \begin{tabular}{l@{\hspace{2mm}}|@{\hspace{2mm}}c}
    \toprule
        Variant & Top-1 Accuracy \\
        \midrule
        Application Prob. + Gumbel-Softmax & 78.36 $\pm$ .51 \\
        Application Prob. + Gumbel-Sinkhorn & 79.11 $\pm$ .30 \\
        FreeAugment & \textbf{80.04 $\pm$ .23}\\
    \bottomrule
    \end{tabular}
\end{table}

\subsection{Continue Evaluation with End-of-Search Classifier}
\cref{tab:ablation-continue-w-search-classifier} compares the results of the evaluation phase where the classifier is being initialized with the end-of-search model weights, with a linear warm-up of 5 epochs for its learning rate, versus training the classifier from scratch. As can be seen from the table, training the classifier from scratch achieves better performance, as it is effectively trained with an optimized augmentation policy from the beginning of its training.

\begin{table}
    \centering
    \caption{Top-1 test accuracy (\%) on CIFAR100 for WRN-40-2 where the evaluation phase continues with the same classifier from the end of the search versus a classifier that was trained from scratch. The results are averaged over 5 random seeds, with the 95\% confidence interval denoted by $\pm$.}
    \label{tab:ablation-continue-w-search-classifier}
    \begin{tabular}{l@{\hspace{2mm}}|@{\hspace{2mm}}c}
    \toprule
        \OurMethod~variant & Top-1 Accuracy \\
        \midrule
        End-of-search & 79.85 $\pm$ .1 \\
        Scratch & \textbf{80.04 $\pm$ .23}\\
    \bottomrule
    \end{tabular}
\end{table}

\newpage
\section{Additional Experiments}
\cref{tab:imagenet1k-results} shows the performance of FreeAugment on the complete ImageNet~\cite{ImageNet} dataset. While evaluating with ResNet-50, we used the found augmentation policy on ImageNet-100 for ResNet-18. FreeAugment surpasses the baseline by \textbf{+1.33\%} and achieves competitive performance when compared to more recent methods.

\begin{table*}[h]
\centering
\caption{Top-1 test accuracy (\%) on ImageNet for ResNet-50 using the found augmentation policy on ImageNet-100. Results are obtained following TA training recipe and averaged across 4 random seeds, with the 95\% confidence interval denoted by $\pm$.}
\label{tab:imagenet1k-results}
\scriptsize
\resizebox{1.0\textwidth}{!}
{
\begin{tabular}{@{}lccccccccc@{}}
\toprule
\multicolumn{1}{l}{\textbf{}} & \multicolumn{1}{c}{Baseline} & 
\multicolumn{1}{c}{AA} &
\multicolumn{1}{c}{FastAA} &
\multicolumn{1}{c}{FasterAA} &
\multicolumn{1}{c}{DADA} &
\multicolumn{1}{c}{TA(Wide)} &
\multicolumn{1}{c}{DDAS} &
\multicolumn{1}{c}{DRA}  &
\multicolumn{1}{c}{\OurMethod}
\\ 
\midrule 
ResNet-50 & 76.3 & 77.6 & 77.6 & 76.5 & 77.5 & 78.07& 77.7& 78.19& 77.63 $\pm$ .15 \\
\bottomrule
\end{tabular}
}
\label{tab:shakeshake}
\vspace{-3mm}
\end{table*}

\cref{tab:domainnet_results_slack} and \ref{tab:imagenet_results_slack} present the same experiments as of 
Tab. 3 and 2 in the main paper,
respectively, with the only change of replacing the more general training scheme proposed by TA~\cite{TrivialAugment} by a tailored one for SLACK~\cite{Slack}. 
\OurMethod~shows competitive results even when adopting SLACK's designated training scheme.

\begin{table*}[h!]
  \centering
  \caption{Top-1 test accuracy (\%) on DomainNet for ResNet-18. The left value in each cell is obtained using SLACK's training recipe, while the right value is a reproduced result using TA's training recipe. All results are averaged across 4 random seeds.}
  \label{tab:domainnet_results_slack}
  \begin{tabular}{l@{\hspace{2mm}}|@{\hspace{2mm}}c@{\hspace{4mm}}c@{\hspace{4mm}}c@{\hspace{4mm}}l}
    \toprule
    & DomainBed & TA (Wide) & SLACK &\OurMethod\\
    \midrule
    Real       & 62.54, 61.91  & 71.56, 70.27 & 71.00, 68.94 & 71.14, 71.47 \\
    Quickdraw  & 66.54, 64.55  & 68.60, 67.01 & 68.14, 66.39 & 68.58, 68.62 \\
    Infograph  & 26.76, 25.64  & 35.44, 33.73 & 34.78, 30.92 & 34.72, 34.88 \\
    Sketch     & 59.54, 58.57  & 66.21, 65.38 & 65.41, 63.75 & 66.48, 66.74 \\
    Painting   & 58.31, 57.70  & 65.15, 64.13 & 64.83, 62.26 & 64.30, 64.65 \\
    Clipart    & 66.23, 64.07  & 71.19, 69.77 & 72.65, 70.92 & 71.06, 71.26 \\
    \midrule
    Average    & 57.23, 55.40  & 63.03, 61.71 & 62.80, 60.53 & 62.71, 62.93 \\
    \bottomrule
  \end{tabular}
\end{table*}

\begin{table}
    \centering
    \caption{Top-1 test accuracy (\%) on ImageNet-100 for ResNet-18. The left value in each cell is obtained using SLACK's training recipe, while the right value is a reproduced result using TA's training recipe. All results are averaged across 4 random seeds.}
    \label{tab:imagenet_results_slack}
    \begin{tabular}{c@{\hspace{4mm}}c@{\hspace{4mm}}c@{\hspace{4mm}}c@{\hspace{4mm}}c@{\hspace{4mm}}c}
    \toprule
         & Baseline & TA(Wide) & SLACK & \OurMethod \\
         \midrule
        ResNet-18 & -, 84.84 & 86.39, 85.68 & 86.06, 86.19 & 86.26, 86.62 \\
    \bottomrule \\
    \end{tabular}
\end{table}

\section{Learning Magnitudes of Non-Differentiable Transformations}

It is important to note that some transformations are associated with a learnable magnitude but are not differentiable with respect to it (e.g., posterize and solarize). To get gradients for such transformations' magnitudes, we use the straight-trough estimator as:

\begin{gather}
    \label{eq:magnitude-straight-through}
    \frac{\partial \SingleImage^{k+1}}{\partial M_{ik}} = \frac{\partial \SingleOp_i(\SingleImage^{k})}{\partial M_{ik}} = \mathbf{1}
\end{gather}

where $\SingleImage^{k}$, is the input image of the $k^\text{th}$ policy layer, $\SingleImage^{k+1}$ is the output image of the same layer, $\SingleOp_i$ is a non-differentiable elementary transformation, $M_{ik}$ is the sampled magnitude at the $k^\text{th}$ policy layer for $\SingleOp_i$, and $\mathbf{1}$ is a matrix from the same size as $\SingleImage^{k}$ filled with ones.
Namely, \cref{eq:magnitude-straight-through} means that the gradient of each pixel in the output image $\SingleImage^{k+1}$ w.r.t. the sampled magnitude $\SingleImage^{k+1}$ equals 1.


In practice, this trick can be easily implemented via:
\begin{gather}
    \label{eq:magnitude-straight-through-implementation}
    \hat{\SingleImage}^{k+1}=\SingleImage^{k+1} + M_{ik} - \text{StopGrad}(M_{ik}),
\end{gather}
where $\text{StopGrad}(\cdot)$ is the analog to PyTorch's detach function, which returns the input value without passing its gradient.
Note that 
\( \partial \SingleImage^{k+1} / \partial M_{ik} = \mathbf{0} \),
where $\mathbf{0}$ is a matrix with the same size of $\SingleImage^{k+1}$ filled with zeros. Thus, the use of the trick in \cref{eq:magnitude-straight-through-implementation} eventually makes \cref{eq:magnitude-straight-through} hold.

\begin{figure}[p]
\begin{lstlisting}[language=Python]
import torch
from torch.nn.functional import gumbel_softmax
from ops import gumbel_sinkhorn, uniform
from torch.distributions.uniform import 

def FreeAugment(x, trans, phi):

    delta, Pi, mu = phi

    N, K = Pi.shape
    
    assert len(trans) == N
    assert len(delta) == K+1
    assert mu.shape[:-1] == Pi.shape
    
    bs = x.shape[0]

    delta = delta.repeat(bs, 1) # shape: (bs, K+1)
    Pi = Pi.repeat(bs, 1, 1) # shape: (bs, N, K)
    mu = mu.repeat(bs, 1, 1, 1) # shape: (bs, N, K, 2)
    
    d_hard = gumbel_softmax(delta) # shape: (bs, K+1)
    P_hard = gumbel_sinkhorn(Pi) # shape: (bs, N, K)
    M = uniform(mu) # shape: (bs, N, K)

    d_hard = d_hard.permute(1,0) # shape: (K+1, bs)
    P_hard = P_hard.permute(2,1,0) # shape: (K, N, bs)
    M = M.permute(2,1,0) # shape: (K, N, bs)

    out = d_hard[0] * x 
    for k, (dh, Ph) in enumerate(zip(d_hard[1:], P_hard)):
        for i, (t_oh, t) in enumerate(zip(Ph, trans)):
            out += dh.view(-1, 1, 1, 1) * \
                   t_oh.view(-1, 1, 1, 1) * t(x, M[k,i])
    
    return out
\end{lstlisting}
\caption{PyTorch per-image policy sampling code}\label{code:per-image-sampling-code}
\end{figure}

\end{document}